\documentclass[10pt,twocolumn,letterpaper]{article}

\usepackage[pagenumbers]{iccv} 

\definecolor{iccvblue}{rgb}{0.21,0.49,0.74}
\usepackage[pagebackref,breaklinks,colorlinks,allcolors=iccvblue]{hyperref}

\usepackage{graphicx}
\usepackage{amsmath}
\usepackage{amssymb}
\usepackage{booktabs}
\usepackage{float}
\usepackage[export]{adjustbox}
\usepackage{subcaption}
\usepackage{caption}
\usepackage{tabularx}
\usepackage{multirow}
\usepackage{array}
\usepackage{rotating}
\usepackage{helvet}
\usepackage{enumitem}
\usepackage{makecell}

\def\paperID{8510} 
\def\confName{ICCV}
\def\confYear{2025}

\title{Beyond RGB: Adaptive Parallel Processing for RAW Object Detection}

\author{
    Shani Gamrian$^{1}$ \quad Hila Barel$^{1}$ \quad Feiran Li$^{1}$ \quad Masakazu Yoshimura$^{2}$ \quad Daisuke Iso$^{1}$ \\
    $^1$Institution A \quad $^2$Institution B
}

\author{
    Shani Gamrian$^{1}$ \quad Hila Barel$^{1}$ \quad Feiran Li$^{1}$ \quad Masakazu Yoshimura$^{2}$ \quad Daisuke Iso$^{1}$ \\
    $^1$Sony Research \quad $^2$Sony Group Corporation \\
    {\tt\small \{shani.gamrian, hila.barel, feiran.li, masakazu.yoshimura, daisuke.iso\}@sony.com}
}

\begin{document}
\maketitle
\begin{abstract}
Object detection models are typically applied to standard RGB images processed through Image Signal Processing (ISP) pipelines, which are designed to enhance sensor-captured RAW images for human vision. However, these ISP functions can lead to a loss of critical information that may be essential in optimizing for computer vision tasks, such as object detection. In this work, we introduce Raw Adaptation Module (RAM), a module designed to replace the traditional ISP, with parameters optimized specifically for RAW object detection. Inspired by the parallel processing mechanisms of the human visual system, RAM departs from existing learned ISP methods by applying multiple ISP functions in parallel rather than sequentially, allowing for a more comprehensive capture of image features. These processed representations are then fused in a specialized module, which dynamically integrates and optimizes the information for the target task. This novel approach not only leverages the full potential of RAW sensor data but also enables task-specific pre-processing, resulting in superior object detection performance. Our approach outperforms RGB-based methods and achieves state-of-the-art results across diverse RAW image datasets under varying lighting conditions and dynamic ranges. Our code is available at \url{https://github.com/SonyResearch/RawAdaptationModule}.
\end{abstract}    
\section{Introduction}
\label{sec:intro}

Most computer vision tasks today are trained and evaluated on standard RGB (sRGB) data, largely due to the popularity of public sRGB datasets like COCO \cite{lin2014microsoft} and ImageNet \cite{deng2009imagenet}. The Image Signal Processing (ISP) pipeline, which converts RAW sensor data into an 8-bit sRGB image, was designed to optimize images for human vision, focusing on visual quality enhancements. This process involves transformations including demosaicing, white balance, noise reduction, color correction, tone mapping, and sharpening, applied sequentially.  While these operations improve aesthetics, they also introduce irreversible information loss. For instance, demosaicing can blur fine patterns, noise reduction may suppress textures, and tone mapping compresses dynamic range, diminishing critical details in highlights and shadows. Furthermore, compression formats like JPEG further increase the loss by discarding data to save space.

To mitigate these issues, recent research has explored the use of RAW images for low-level vision tasks such as super-resolution, deblurring, and denoising \cite{lecouat2022high, liang2020raw, kerepecky2021d3net, wang2020practical, zhang2021rethinking}. More recently, RAW data has been leveraged for high-level tasks like object detection, classification, and segmentation \cite{yu2021reconfigisp, xu2023toward, Yoshimura_2023_ICCV, maxwell2024logarithmic, chen2023instance}, as it retains full sensor information and improves performance in challenging conditions such as low-light and noisy environments \cite{Chen_Tai_Ma_2024, morawski2022genisp}. However, training deep neural networks directly on RAW data presents challenges. RAW pixels, particularly in high-dynamic-range (HDR) images, are often heavily skewed toward zero, with a sparse distribution of highlights, making it difficult for networks to detect essential features such as edges and patterns. Therefore, rather than removing the ISP entirely, we propose replacing it with an adaptive component that processes RAW input into a task-optimized representation.


    

\begin{figure}
    \centering
    
    \includegraphics[trim=105 280 50 80, clip, width=\linewidth]{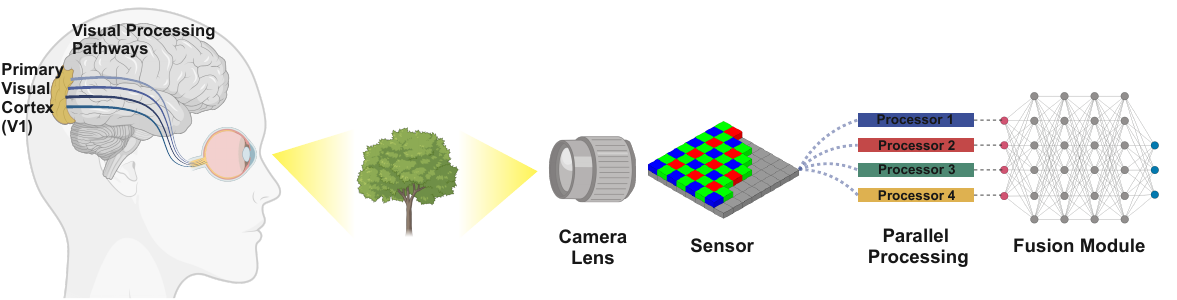} 
    \caption{A simplified illustration of the human visual pathways (left) and our parallel pipeline (right), where multiple ISP operations process the input simultaneously before fusion, inspired by the parallel processing in the human visual system.}
    \label{fig:brain}
    \vspace{-2mm}
\end{figure}

Existing ISP pipelines, traditional and learned, follow a sequential design, where each stage modifies the output of the previous one. This approach can be suboptimal, as each transformation is influenced by the cumulative effects of earlier stages. As a result, crucial information may be lost or altered before reaching the final stages, limiting the object detection network’s ability to model complex interactions. For example, white balance applies channel-wise gains early in the ISP, potentially clipping near-saturated areas and permanently losing highlight details. This limits later stages, affecting final image quality and object detection. To address this, we draw inspiration from the human visual system and propose replacing the conventional sequential pipeline with a learned, adaptive parallel approach.

As illustrated in \cref{fig:brain} (left), the human retina processes visual information through parallel pathways that extract features like color, contrast, and fine details, which are later integrated and further processed in the primary visual cortex (V1) for higher-level interpretation \cite{nassi2009parallel}. Drawing inspiration from this parallel processing behavior, we introduce the Raw Adaptation Module (RAM), a novel pre-processing module that operates directly on RAW images. Instead of applying ISP operations sequentially, RAM processes multiple attributes in parallel and fuses them into a unified representation, similar to the integrative role of V1.

RAM acts as a pre-processing stage, converting RAW images into a representation fed into the backbone and detector, with the entire pipeline trained end-to-end and optimized solely through the object detection loss. Due to its flexible design, RAM can be integrated with any backbone and detector, making it highly adaptable for various pipelines. Additionally, RAM is efficient in terms of parameters and latency, making it suitable for high and low-computation hardware. While prior works have shown improvements on individual datasets with similar characteristics \cite{10.1007/978-3-031-31435-3_25, morawski2022genisp, xu2023toward}, our work presents a more comprehensive approach. To summarize, our main contributions are:

\begin{itemize}[leftmargin=0.75cm] 
\item We present an end-to-end RAW object detection framework that incorporates a novel parallel ISP approach. This method integrates outputs from multiple ISP functions, enabling the model to learn a task-specific representation that optimizes detection performance on RAW images by leveraging full sensor data.
\item We achieve state-of-the-art results across seven public RAW datasets with varying dynamic ranges (12-bit to 24-bit) and challenging conditions, such as low-light, diverse exposures, extreme weather, and sensor noise. Our method outperforms sRGB data and recent works, highlighting the potential of RAW data for vision tasks.
\item We present a Shapley values analysis \cite{f99c1a45-348b-3431-979a-6234c790659b} to highlight the modular architecture of RAM, showing the impact of individual ISP functions on performance and underscoring the benefits of a parallel processing structure. 
\end{itemize}


\section{Related Work}
\label{sec:related_work}
\newcommand{\squeezeup}{\vspace{-1.0mm}}


\begin{figure*}[t]
    \centering
    \includegraphics[width=0.99\textwidth]{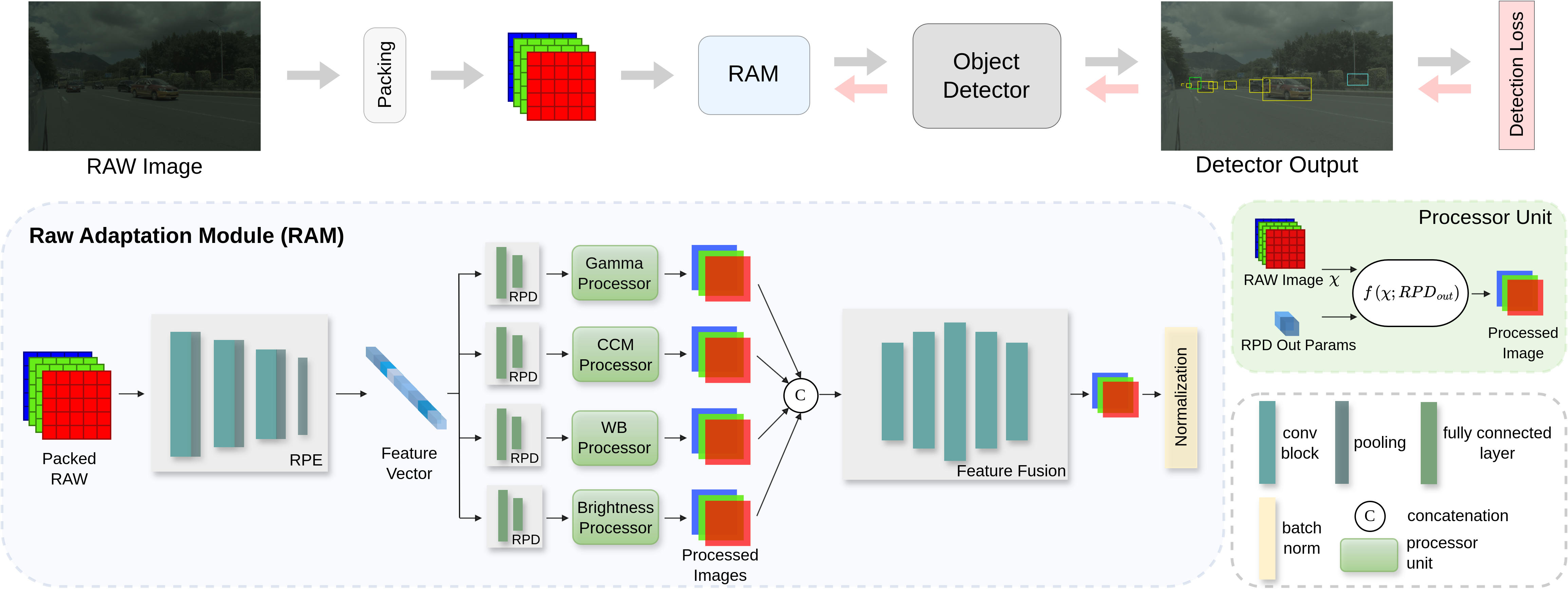} 
    \caption{Overview of our proposed end-to-end RAW object detection pipeline with RAM: The input is first reshaped into an RGGB representation and then passed through the RPEncoder to generate a shared feature vector for multiple RPDecoders and Processors, each optimizing and applying a specific ISP function on the input image. These outputs are fused in the Feature Fusion module to create an optimized representation, fed into the object detector. The entire pipeline is jointly trained using the detection losses.}
    \label{fig:ram}
\end{figure*}

\subsection{Object Detection}

Object detection has advanced significantly in recent years, with approaches generally categorized into two main types: two-stage and one-stage detectors. Two-stage detectors, such as Faster R-CNN \cite{conf/nips/RenHGS15} and Cascade R-CNN \cite{Cai2017CascadeRD}, first propose regions of interest and then classify these regions, offering high accuracy at the cost of computational complexity. In contrast, one-stage detectors like SSD \cite{10.1007/978-3-319-46448-0_2}, RetinaNet \cite{8237586}, and the YOLO series \cite{7780460, ge2021yolox} perform detection and classification simultaneously, prioritizing speed and efficiency. Transformers, first introduced for natural language processing \cite{vaswani2017attention}, have revolutionized computer vision tasks, including object detection. DETR (DEtection TRansformer) \cite{10.1007/978-3-030-58452-8_13} introduced an encoder-decoder architecture with cross-attention mechanisms, eliminating the need for hand-designed components like non-maximum suppression (NMS). Despite the innovative approach of DETR, its slow convergence made it less practical for common use. Subsequent works such as DN-DETR \cite{li2022dn} and DINO \cite{zhang2022dino} have addressed these limitations through denoising training, anchor initialization, and refined query selection. Our proposed RAM pipeline is designed to be independent of the object detector architecture and can be integrated with any detector. For a comprehensive evaluation, we employ both the convolutional-based Faster R-CNN, YOLOX and the transformer-based DINO, showcasing the flexibility of our approach across different detection architectures.

\subsection{RAW Object Detection}
RAW images have the potential to improve object detection beyond sRGB. However, recent studies indicate that DNNs often struggle with RAW data distribution \cite{hansen2021isp4ml, yoshimura2024pqdynamicisp}, leading to decreased accuracy \cite{yoshimura2023rawgment, 10.1007/978-3-031-31435-3_25, xu2023toward}. Three approaches have been adopted to enhance RAW object detection. The first approach incorporates ISP processing into the RAW detector via distillation. ISP Distillation \cite{schwartz2023isp} aligns the features of the RAW detector with those of an sRGB detector, while LOD \cite{Hong2021Crafting} adds a reconstruction loss to restore clean sRGB images from low-light RAW inputs. The second approach tunes the hyper-parameters of ISPs to produce more recognizable images for a pre-trained sRGB object detector, addressing the domain gaps caused by the difference of image sensors used in training and inference. Evolutionary algorithms \cite{mosleh2020hardware} or gradients from a detection loss \cite{tseng2019hyperparameter} are used to tune. Recent improvements focus on controlling ISP parameters per image rather than tuning them \cite{qin2023learning, sun2024rl, qin2022attention}. The third approach trains the ISP and detector end-to-end, optimizing both with object detection loss \cite{yoshimura2023rawgment, 10.1007/978-3-031-31435-3_25, onzon2021neural, cui2025raw}. DynamicISP \cite{Yoshimura_2023_ICCV} and AdaptiveISP \cite{wang2024adaptiveisp} improve accuracy by dynamically controlling ISP parameters, while GenISP \cite{morawski2022genisp} and ROAD \cite{xu2023toward} generates them using a CNN.

While these approaches significantly improve accuracy without being constrained by the sRGB format, they rely on classical ISP piplines that apply ISP functions in series, which can be suboptimal. In fact, we found that some ISP functions may be harmful for certain datasets, which is why we propose a parallel pipeline where our method receives outputs from different functions and creates an output that is optimized for the specific data and task.

\section{Raw Adaptation Module}
\label{sec:ram}

\subsection{Overview}
Existing state-of-the-art object detectors are trained and evaluated on sRGB images processed with ISP parameters calibrated for human vision. Unlike traditional ISPs, which apply fixed functions in sequence, RAM introduces a novel approach where multiple ISP functions are applied in parallel. Parallelism is a fundamental characteristic of the human visual system, where multiple visual pathways operate simultaneously. This parallel processing mechanism enables rapid scene analysis and efficient extraction of critical details. Drawing inspiration from this, RAM applies multiple ISP functions independently to the RAW image, preserving more information and enabling richer feature combinations. It dynamically selects and weighs the most relevant features for each image, discarding unnecessary or potentially harmful functions. This ensures that only task-optimized transformations contribute to the performance.

Our pipeline begins by converting Bayer pattern RAW images into an RGGB-stacked format. The input is then fed into RAM, which produces a processed image tailored to the input data while maintaining the same spatial dimensions. This processed image is then passed to the backbone and object detector, which performs detection on the optimized input representation. The following sections provide an overview of the core components of our architecture.

\subsection{Module Design}
Our approach is designed to adaptively optimize the input representation for object detection, refining ISP parameters and fusing relevant features into a unified output, as illustrated in \cref{fig:ram}. The module consists of two main components: Adaptive ISP Parameterization and Feature Fusion.
\subsubsection{Adaptive ISP Parameterization}
The Adaptive ISP Parameterization component of RAM is responsible for predicting and applying the ISP parameters adaptively. RAM employs a Raw Parameter Encoder (RPE) and multiple Raw Parameter Decoders (RPD). The RPE is a convolutional-based encoder that takes the RAW input image and generates a feature vector. The RPD is an MLP decoder that takes the feature vector and predicts the corresponding ISP function parameters.

In the RAM architecture, a single RPE is applied to the RAW input image, and the resulting feature vector is shared among separate RPDs for each ISP function. This approach is motivated by the observation that many ISP functions can benefit from similar input features, making it unnecessary to extract separate feature vectors for each function. Each processor then applies the corresponding ISP function to the input image in parallel using the predicted parameters. This parallel architecture is crucial not only for efficiency but also for ensuring that all potential features are available to the subsequent Feature Fusion module.


The ISP functions incorporated into RAM are selected based on their relevance to our datasets. These include gamma correction, brightness adjustment, color correction, and white balancing.

Gamma correction, used for tone mapping, adjusts the input image $\mathcal{X}$ as follows:
\begin{equation}
\mathcal{X}_{\textit{gamma}} = \mathcal{X}^{\mathcal{RPD}(\mathcal{RPE(X)}; \theta_{\textit{gamma}})}
\end{equation}
which is particularly important for high-dynamic range datasets. Additionally, brightness adjustment is applied by adding a predicted offset to the input image:
\begin{equation}
\mathcal{X}_{\textit{brightness}} = \mathcal{X} + \mathcal{RPD}(\mathcal{RPE(X)}; \theta_{\textit{brightness}})
\end{equation}

The color correction matrix (CCM) modifies the input RGB channels by multiplying a learned $3 \times 3$ matrix:
\begin{equation}
\mathcal{X}_{\textit{ccm}} = 
\begin{bmatrix}
  c_{11} & c_{12} & c_{13} \\
  c_{21} & c_{22} & c_{23} \\
  c_{31} & c_{32} & c_{33}
\end{bmatrix}
\cdot
\begin{bmatrix}
  R \\
  G \\
  B
\end{bmatrix}
\end{equation}
where the matrix elements $c_{ij}$ are generated by $\mathcal{RPD}(\mathcal{RPE(X)}; \theta_{\textit{ccm}})$.

For white balance (WB), learned gains for the R, G, and B channels are applied:
\begin{equation}
\mathcal{X}_{\textit{wb}} = 
\begin{bmatrix}
  \alpha_R \\
  \alpha_G \\
  \alpha_B
\end{bmatrix}
\odot
\begin{bmatrix}
  R \\
  G \\
  B
\end{bmatrix}
\end{equation}
where the gains $\alpha$ are generated using $\mathcal{RPD}(\mathcal{RPE(X)}; \theta_{\textit{wb}})$ in a similar way. 


\subsubsection{Feature Fusion} 
To fuse the information from the various processed inputs, the RAM architecture employs a reverse-hourglass design. Initially, all processed inputs are concatenated into a single multi-channel input:
\begin{equation}
\mathcal{X}_{\textit{fused}} = \left[\mathcal{X}_{\textit{gamma}}, \mathcal{X}_{\textit{brightness}}, \mathcal{X}_{\textit{ccm}}, \mathcal{X}_{\textit{wb}}\right]
\end{equation}

\noindent The input is then passed to the Feature Fusion module, where a reverse-hourglass architecture combines information from different inputs to generate a unified, optimized 3-channel representation with the same spatial dimensions as the original input. The structure features a wider middle layer that emphasizes feature richness, enabling the model to capture both detailed and contextual information before narrowing back down, resulting in an optimized representation that retains critical features. This module selectively emphasizes relevant features for object detection, filtering out less useful or irrelevant ones.

Typical object detection methods use mean-std normalization to standardize data and improve convergence. However, due to the dynamic nature of RAM, we cannot pre-compute the mean and standard deviation (std) for the data. Therefore, we apply batch normalization \cite{ioffe2015batch} after the Feature Fusion step instead, allowing the model to learn appropriate normalization parameters during training. The normalized output is then fed as input to the subsequent object detection model, agnostic to the specific architecture or backbone. 

Incorporating RAM into object detection enables a pipeline that utilizes end-to-end optimization of ISP parameters and detection algorithms, creating a task-specific input representation. This approach enhances performance while improving efficiency through shared feature encodings, resulting in a more adaptable system that challenges traditional RAW-to-RGB pipelines.

Further details on the model architecture are available in the supplementary material.
\section{Experiments}
\label{sec:experiments}
\subsection{Datasets}

\begin{table*}[!ht]
\renewcommand{\arraystretch}{1.05}
\centering
\caption{Quantitative results and comparisons with state-of-the-art methods across different RAW object detection datasets. Results are reported using mean Average Precision (mAP) and mAP at 50\% IoU (mAP$_\text{50}$). The highest result is highlighted in bold, while the second-highest is marked with an underline.}
\adjustbox{max width=\linewidth}{%
\begin{tabular}{lcccccccccccccc}
\hline
Method & \multicolumn{2}{c}{LOD-Dark} & \multicolumn{2}{c}{LOD-Normal} & \multicolumn{2}{c}{ROD-Day} & \multicolumn{2}{c}{ROD-Night} & \multicolumn{2}{c}{NOD-Nikon} & \multicolumn{2}{c}{NOD-Sony} & \multicolumn{2}{c}{PASCALRAW} \\
& mAP & mAP$_\text{50}$ & mAP & mAP$_\text{50}$ & mAP & mAP$_\text{50}$ & mAP & mAP$_\text{50}$ & mAP & mAP$_\text{50}$ & mAP & mAP$_\text{50}$ & mAP & mAP$_\text{50}$ \\
\hline
RAW & 28.5 & 50.8 & 32.7 & 53.8 & 21.4 & 35.7 & 30.4 & 52.1 & 26.4 & 50.3 & 28.7 & 54.2 & 61.5 & 90.0 \\
\hline
sRGB & 28.7 & 51.2 & 34.5 & 57.0 & 24.5 & 40.1 & 38.9 & 62.8 & 27.4 & 52.1 & 28.6 & 53.7 & 64.3 & 92.0 \\
Log RGB \cite{maxwell2024logarithmic} & 32.2 & 54.7 & 35.7 & 57.6 & 23.8 & 39.0 & 39.5 & 63.8 & 28.4 & 52.8 & 27.5 & 51.6 & 64.1 & 91.5 \\
YOLA \cite{hong2024you} & 32.3 & 55.4 & \underline{38.1} & \underline{60.9} & 26.8 & 42.9 & 42.1 & 66.2 & 28.6 & 52.9 & 29.8 & 55.6 & 66.2 & \underline{92.6} \\
FeatEnHancer \cite{hashmi2023featenhancer} & 32.2 & 55.1 & 37.5 & 59.9 & \underline{27.0} & \underline{43.7} & \underline{42.6} & \underline{66.8} & 28.9 & 53.3 & 30.2 & 55.7 & \underline{66.5} & \textbf{92.7} \\
\hline
Gamma~\cite{10.1007/978-3-031-31435-3_25} & 31.5 & 54.1 & 36.3 & 58.7 & 24.5 & 39.8 & 39.6 & 64.0 & 28.0 & 53.0 & 29.0 & 54.0 & 65.0 & 91.9 \\
DynamicISP~\cite{Yoshimura_2023_ICCV} & 32.0 & \underline{56.1} & 37.0 & 60.4 & 25.8 & 42.3 & 41.4 & 65.6 & \underline{30.1} & 54.2 & 30.3 & \underline{55.8} & 62.7 & 91.2 \\
IA-ISPNet \cite{liu2022image} & 32.3 & 54.9 & 36.3 & 58.2 & 25.7 & 41.7 & 40.4 & 64.3 & 27.0 & 52.1 & 28.1 & 53.4 & 64.7 & 91.5 \\
GenISP~\cite{MorawskiCLDHH22} & \underline{32.6} & 55.1 & \underline{38.1} & 60.1 & 24.8 & 39.9 & 40.2 & 64.2 & 29.8 & \underline{54.6} & \underline{30.7} & 55.7 & 64.8 & 91.8 \\
RAOD~\cite{xu2023toward} & 32.3 & 54.8 & 36.5 & 58.0 & 26.2 & 42.9 & 40.6 & 64.7 & 28.9 & 53.8 & 29.1 & 54.2 & 65.3 & 90.9 \\
\hline
\textbf{RAM (Ours)} & \textbf{34.9} & \textbf{57.6} & \textbf{40.1} & \textbf{61.6} & \textbf{28.3} & \textbf{45.1} & \textbf{44.5} & \textbf{69.0} & \textbf{31.0} & \textbf{56.3} & \textbf{32.4} & \textbf{59.1} & \textbf{66.8} & \underline{92.6} \\
\hline
\end{tabular}
}
\label{table_sota}
\end{table*}

To establish the superiority of RAM over traditional ISP (sRGB) and other state-of-the-art methods, we conduct extensive experiments across various RAW object detection datasets.

\textbf{ROD:} The ROD dataset~\cite{xu2023toward} consists of 24-bit HDR RAW images captured during both day and night driving scenes. 
The dataset provides 4,053 daytime images (ROD-Day) and 12,036 nighttime images (ROD-Night), with annotations for five object classes: \textit{tram}, \textit{car}, \textit{truck}, \textit{cyclist}, and \textit{pedestrian}.

\textbf{NOD:} The NOD dataset \cite{MorawskiCLDHH22} includes 14-bit RAW outdoor images taken under low-light conditions. It includes 7,200 images, with 3,200 captured by the Sony RX100 VII (NOD-Sony) and 4,000 by the Nikon D750 (NOD-Nikon). The dataset is annotated with bounding boxes for 46,000 instances of \textit{people}, \textit{bicycle}, and \textit{car} classes.

\textbf{LOD:} The LOD dataset \cite{Hong2021Crafting} contains 2,230 14-bit low-light RAW images categorized into eight object classes. This dataset is designed for detecting multiple categories of objects in low-light indoor and outdoor environments. It includes long (LOD-Normal) and short (LOD-Dark) exposure images of the same scenes. The object classes are: \textit{car}, \textit{motorcycle}, \textit{bicycle}, \textit{chair}, \textit{dining table}, \textit{bottle}, \textit{TV}, and \textit{bus}.

\textbf{PASCALRAW:} The PASCALRAW dataset \cite{omid2014pascalraw} consists of 4,259 daylight 12-bit RAW images, all captured using a Nikon D3200 DSLR camera in daylight conditions across Palo Alto and San Francisco. The dataset includes annotations for instances of \textit{person}, \textit{bicycle}, and \textit{car}.

\subsection{Implementation Details}

Due to resource constraints and the need for faster training and evaluation, we downsampled all dataset inputs to a height of 400 pixels. For the state-of-the-art comparison experiments, we employed the two-stage Faster R-CNN detector \cite{DBLP:journals/corr/RenHG015} with ResNet18 \cite{DBLP:journals/corr/HeZRS15} as the backbone, chosen for its effectiveness on smaller datasets. In contrast, for experiments involving weather conditions and noisy data based on larger datasets, we used the one-stage DINO detector \cite{zhang2023dino} with ResNet50 backbone. Our implementation is based on MMDetection \cite{DBLP:journals/corr/abs-1906-07155}, and all models are trained from scratch and evaluated on NVIDIA H100 GPU. Additional details on the experimental settings can be found in the supplementary material.

\subsection{Comparisons with State-of-the-Art Methods}
\begin{table}
\renewcommand{\arraystretch}{1.0}
\centering
\caption{Comparison of Model Size, Inference Time (measured on NVIDIA H100 GPU), and mAP on the ROD-Night dataset. ‘Baseline’ represents the reference model, with values indicating the additional time cost and parameters introduced by each method.}
\adjustbox{max width=\linewidth}{
\begin{tabular}{lcccc}
\hline
Method & Params (M) & Inference Time (ms) & mAP \\
\hline
RAW (ResNet18) & Baseline (12.3) & Baseline (35.8) & 30.4 \\
RAW (ResNet50) & +12.47 & +2.33 & 34.7 \\
sRGB (ResNet18) & - & - & 38.9 \\
YOLA & +0.03 & +1.43 & 42.1\\
FeatEnHancer & +0.53 & +2.25 & 42.6 \\
Gamma & - & - & 39.6 \\
DynamicISP & +1.46 & +1.3 & 41.4 \\
IA-ISPNet & +0.63 & +1.1 & 40.4 \\
GenISP & +0.57 & +0.86 & 40.2 \\
RAOD & +0.28 & +1.22 & 40.6 \\
\textbf{RAM (Ours)} & +0.54 & +1.05 & \textbf{44.5} \\
\textbf{RAM-T (Ours)} & +0.2 & +1.01 & \underline{44.2} \\
\hline
\end{tabular}
}
\label{table_flops_params}
\vspace{-2mm}
\end{table}

We use RawPy \cite{rawpy_cite}, a widely used open-source ISP pipeline, to convert RAW inputs into sRGB images. Recent methods enhance sRGB for object detection: LogRGB \cite{maxwell2024logarithmic} pre-processes RGB data using logarithmic transformation, while YOLA \cite{hong2024you} and FeatEnHancer \cite{hashmi2023featenhancer} improve feature representation in low light. As shown in \cref{table_sota}, while FeatEnHancer leads among sRGB-based methods, RAM surpasses it by +1.9\% mAP on ROD-Night and +2.7\% on LOD-Dark, while also achieving highest mAP results on the other datasets. These results demonstrate the ability of RAM to replace conventional ISP with a learned alternative that preserves information while surpassing advanced sRGB image enhancement methods.

\begin{figure*}[htbp]
    \centering
    \tiny
    \setlength{\tabcolsep}{-1pt} 
    \begin{tabular}{m{0.01\linewidth}*{7}{>{\centering\arraybackslash}m{0.165\linewidth}}} 
        \rotatebox{90}{\sffamily ROD-Night} &
        \includegraphics[width=0.95\linewidth]{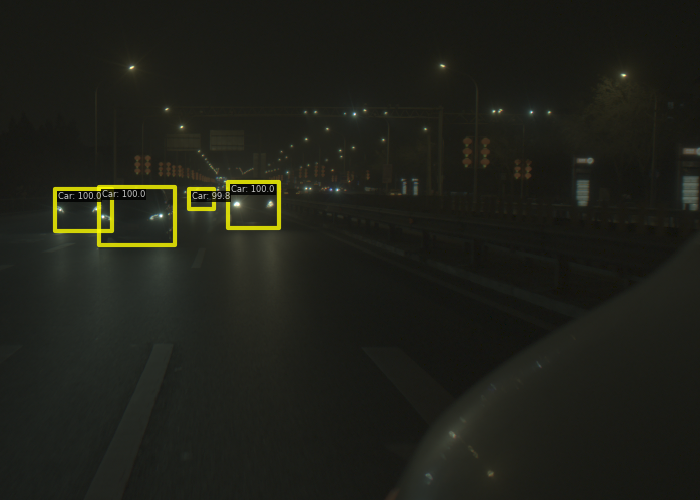} &
        \includegraphics[width=0.95\linewidth]{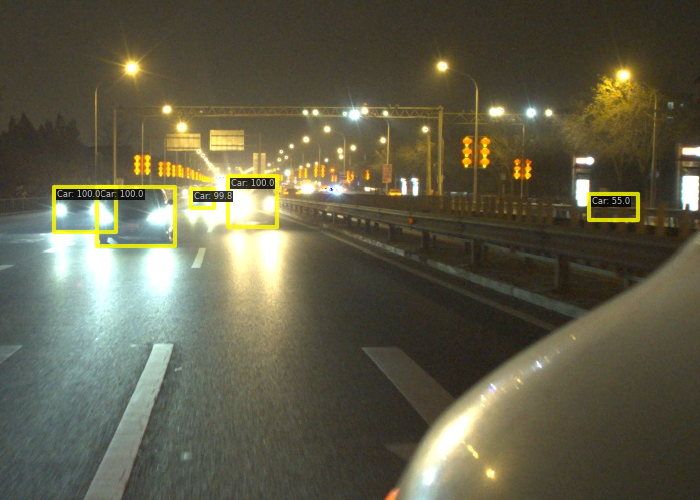} &
        \includegraphics[width=0.95\linewidth]{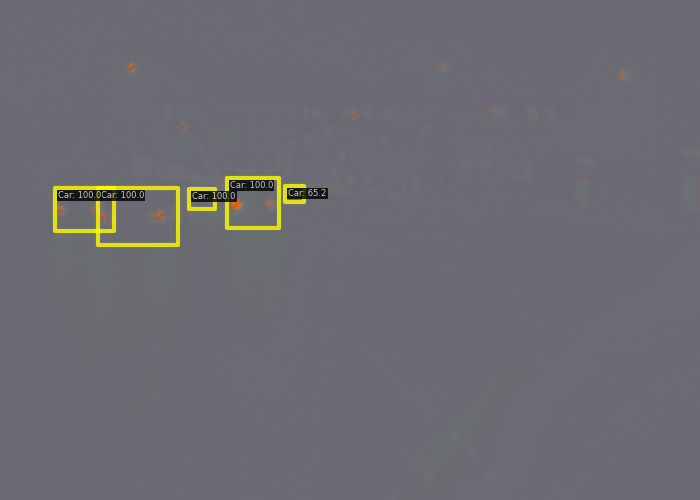} &
        \includegraphics[width=0.95\linewidth]{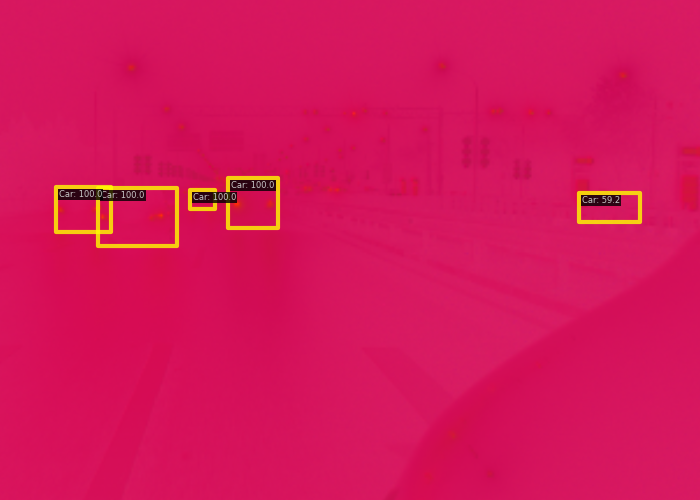} &
        \includegraphics[width=0.95\linewidth]{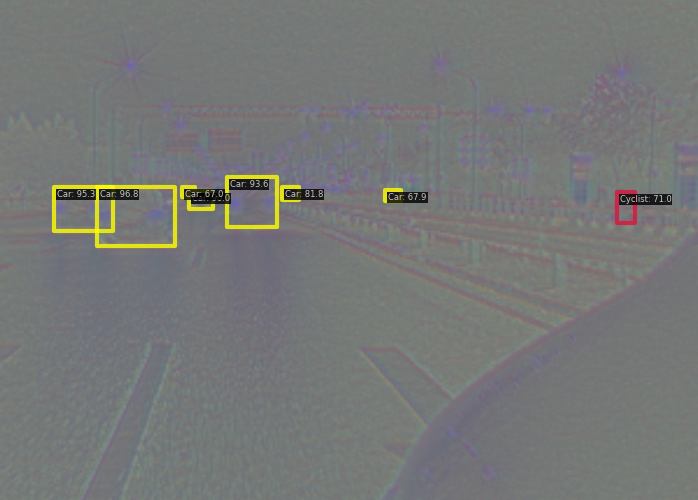} &
        \includegraphics[width=0.95\linewidth]{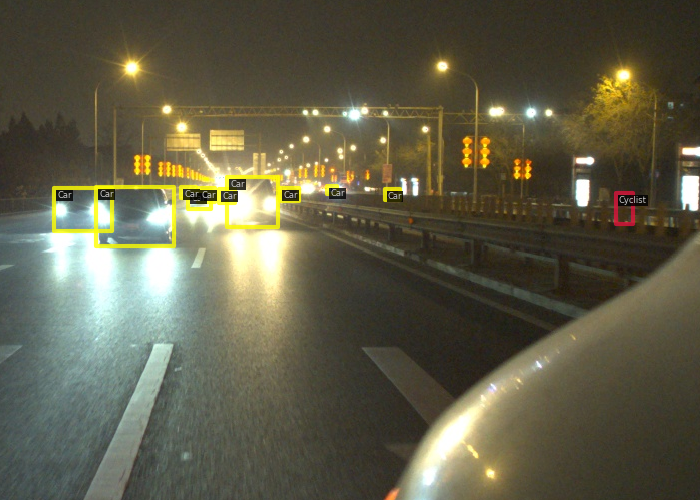}
        \\[2pt]
        
        \rotatebox{90}{\sffamily NOD-Sony} &
        \includegraphics[width=0.95\linewidth]{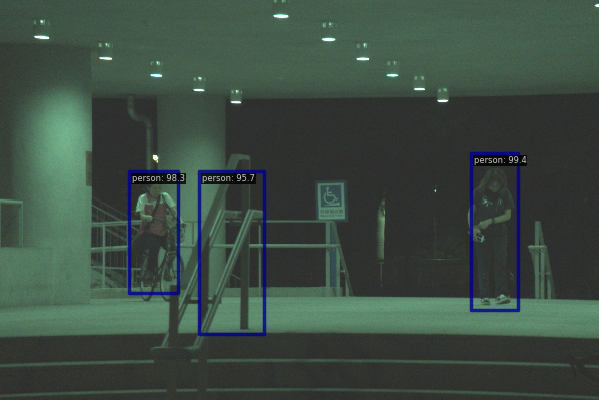} &
        \includegraphics[width=0.95\linewidth]{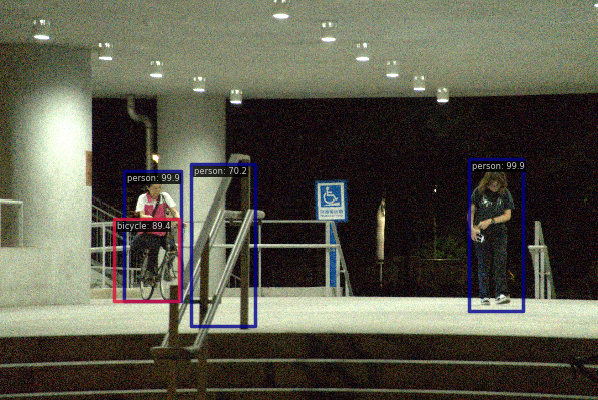} &
        \includegraphics[width=0.95\linewidth]{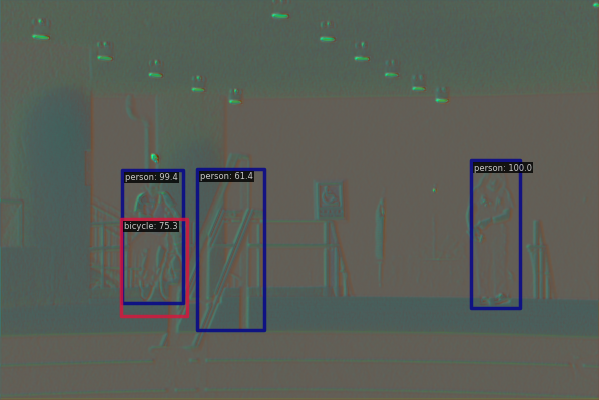} &
        \includegraphics[width=0.95\linewidth]{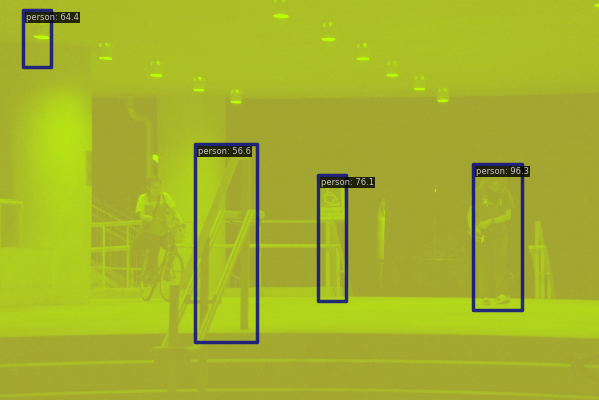} &
        \includegraphics[width=0.95\linewidth]{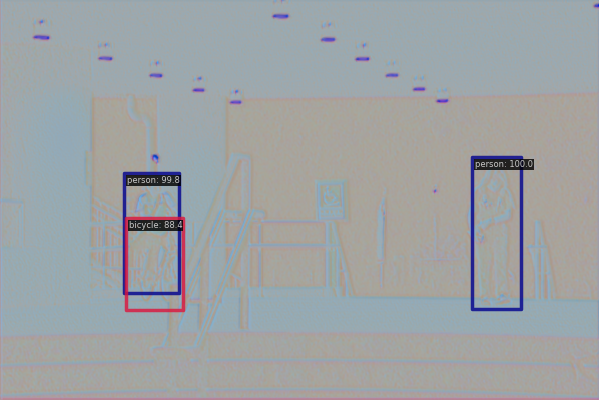} &
        \includegraphics[width=0.95\linewidth]{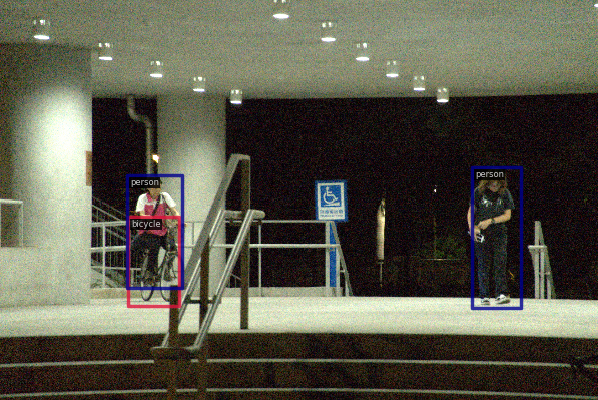}
        \\[2pt]
        
        \rotatebox{90}{\sffamily LOD-Normal} &
        \includegraphics[width=0.95\linewidth]{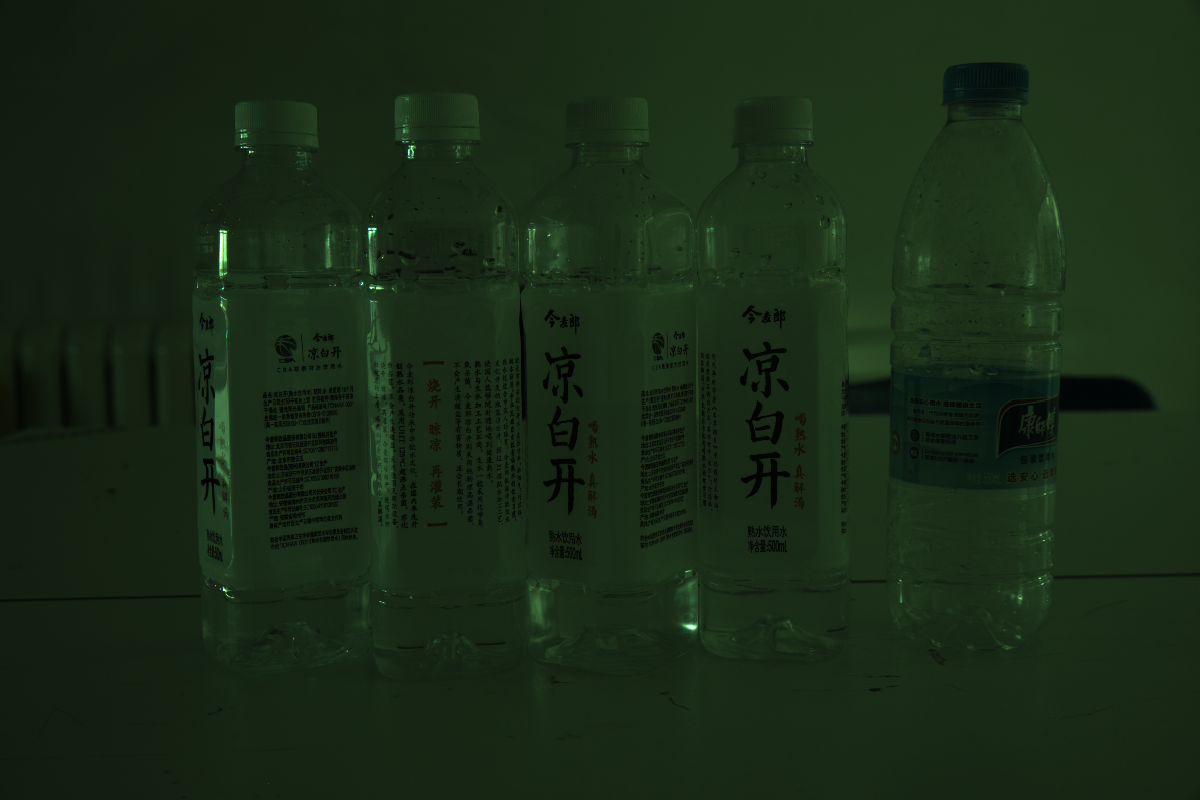} &
        \includegraphics[width=0.95\linewidth]{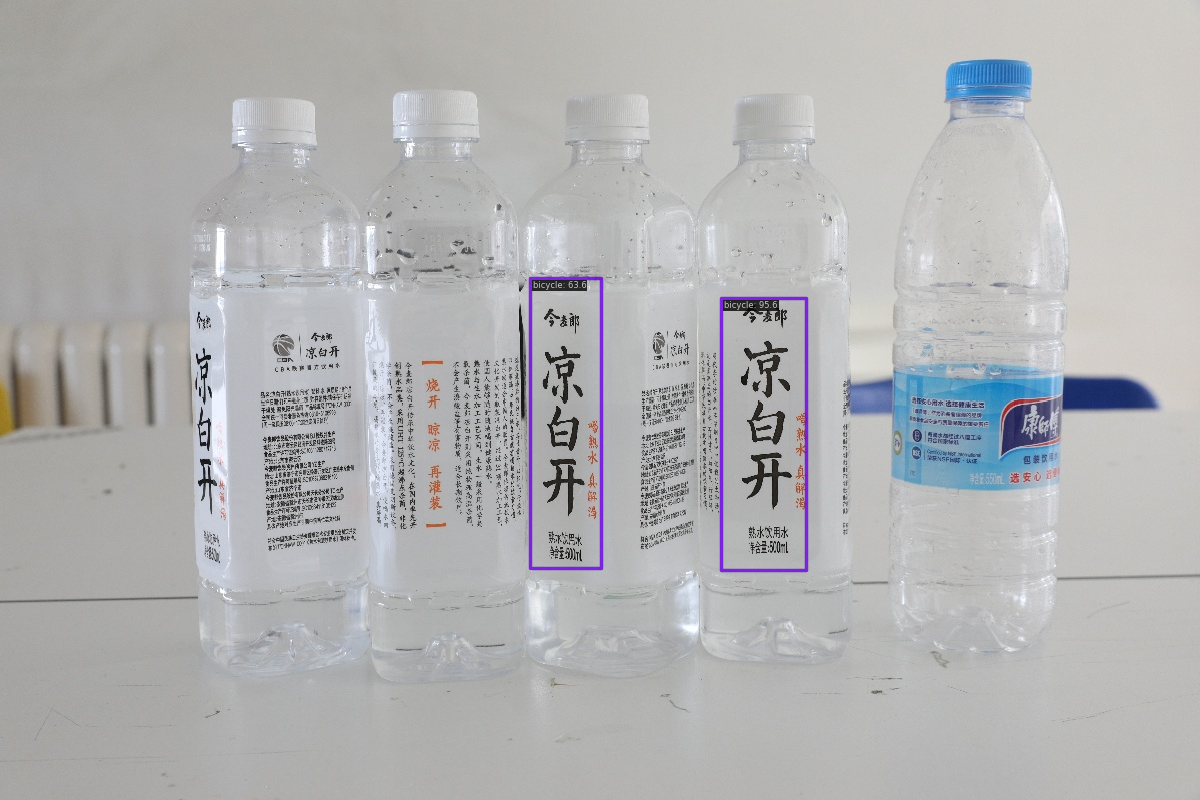} &
        \includegraphics[width=0.95\linewidth]{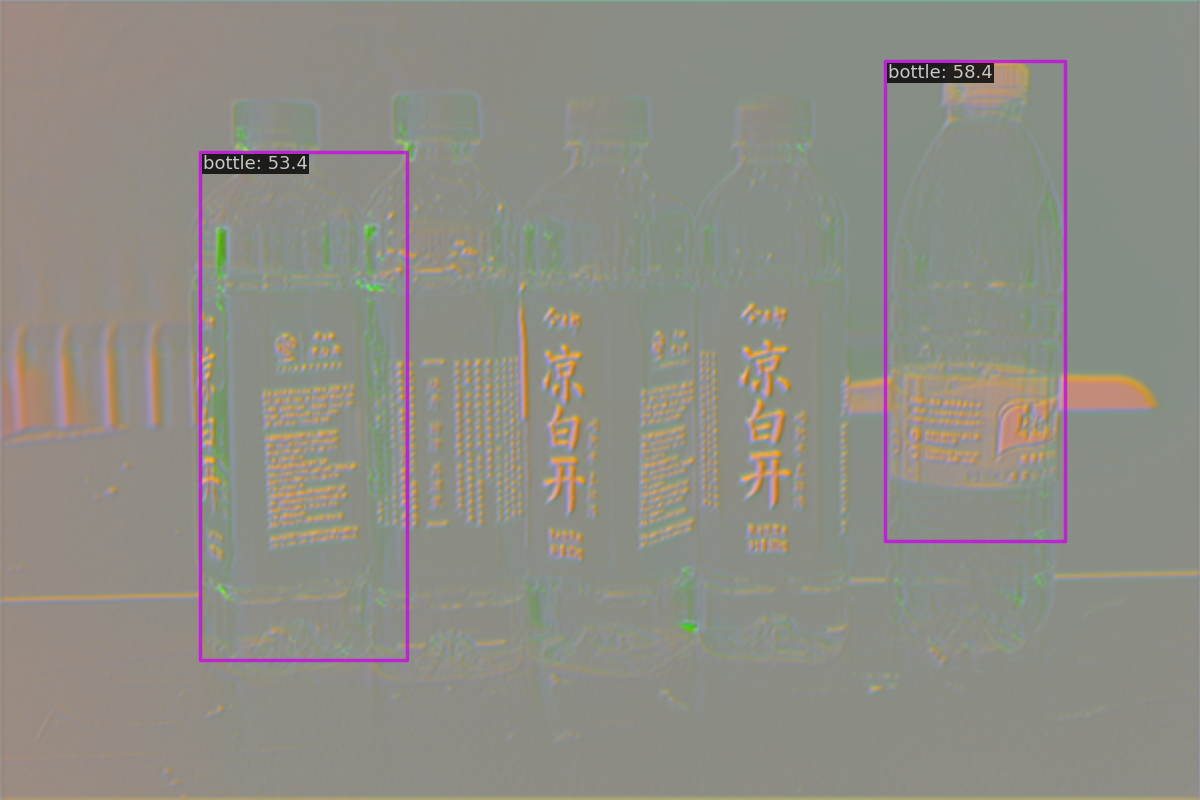} &
        \includegraphics[width=0.95\linewidth]{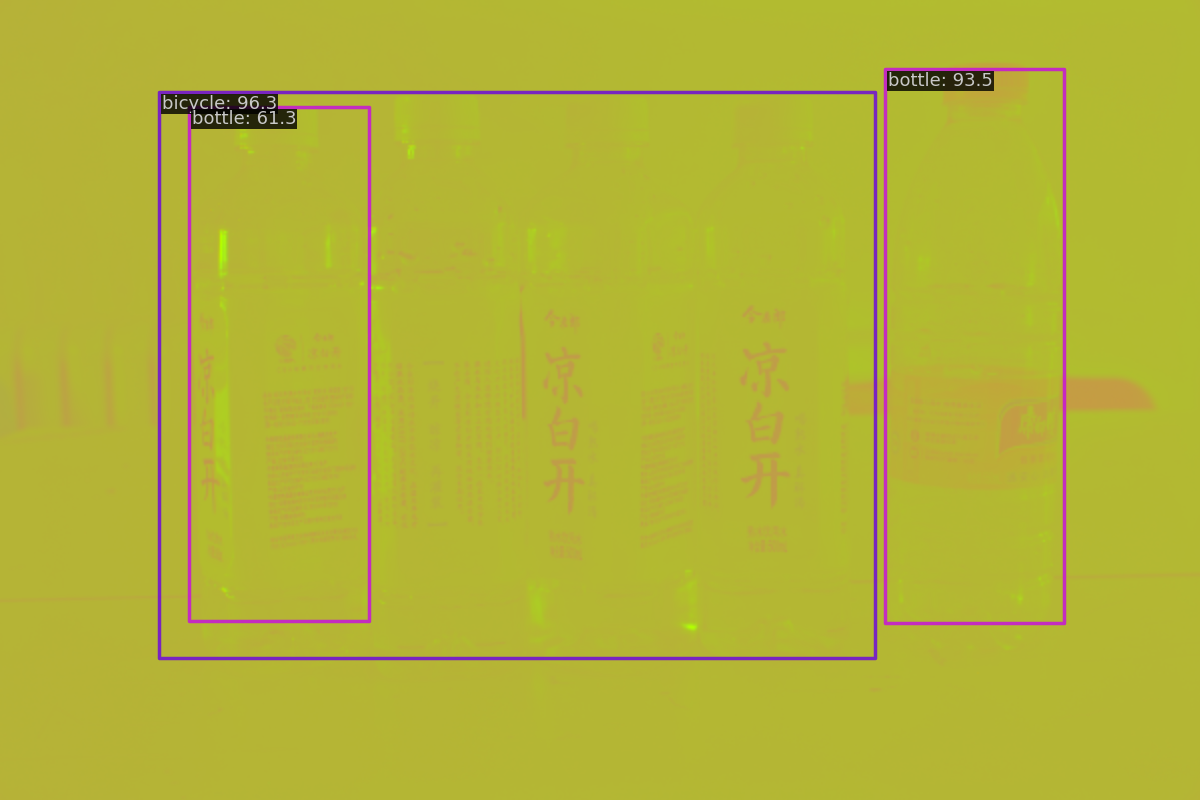} &
        \includegraphics[width=0.95\linewidth]{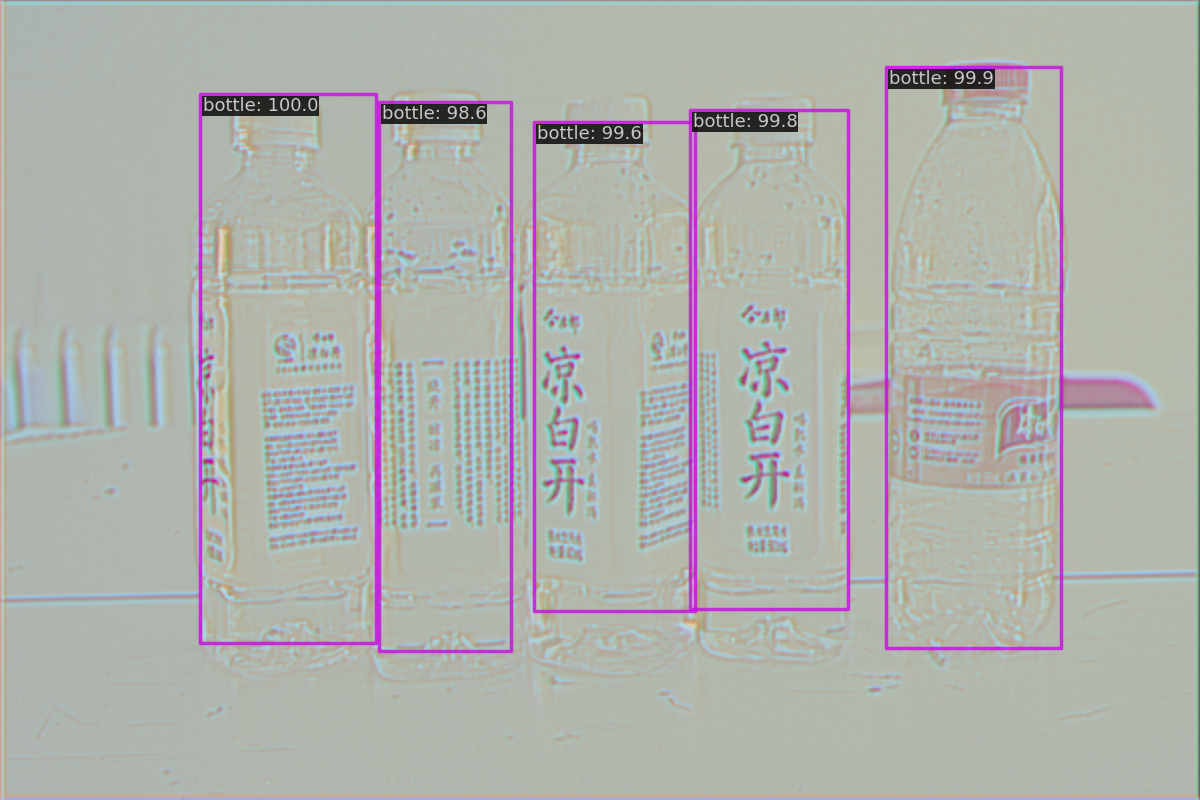} &
        \includegraphics[width=0.95\linewidth]{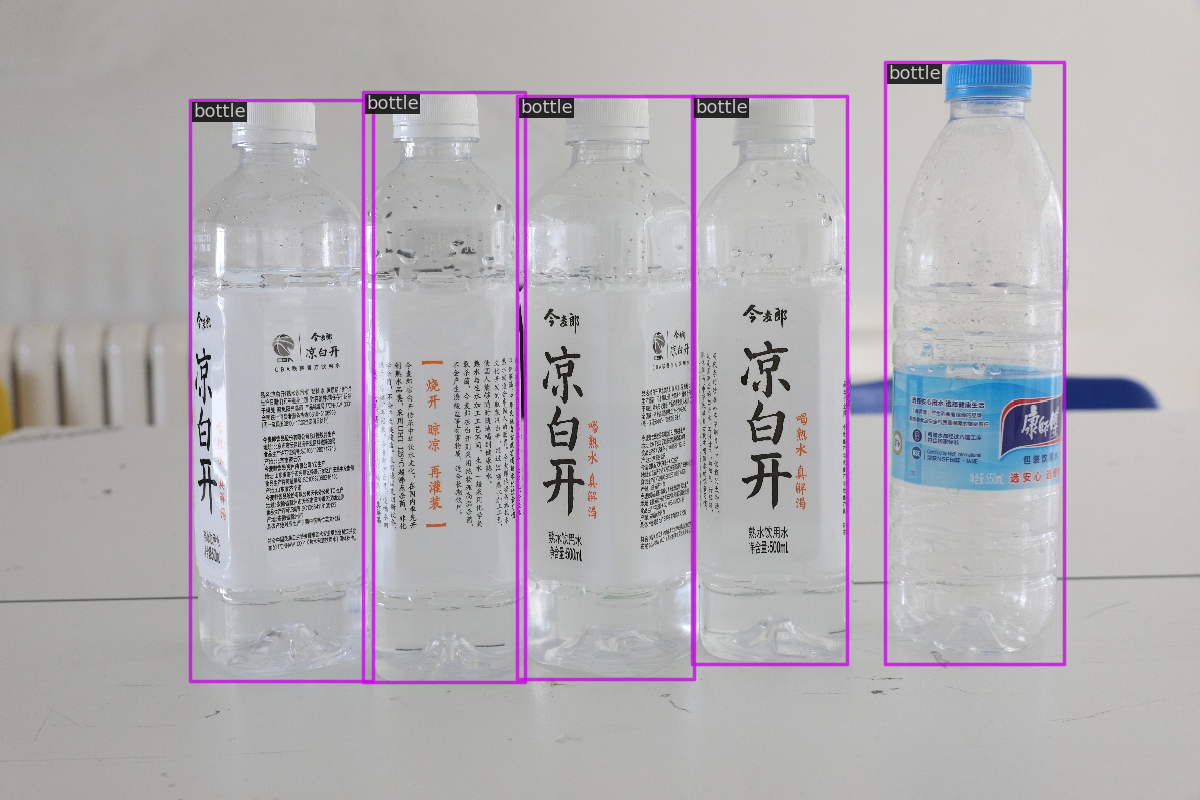}
        \\[2pt]
        
        \rotatebox{90}{\sffamily PASCALRAW} &
        \includegraphics[width=0.95\linewidth]{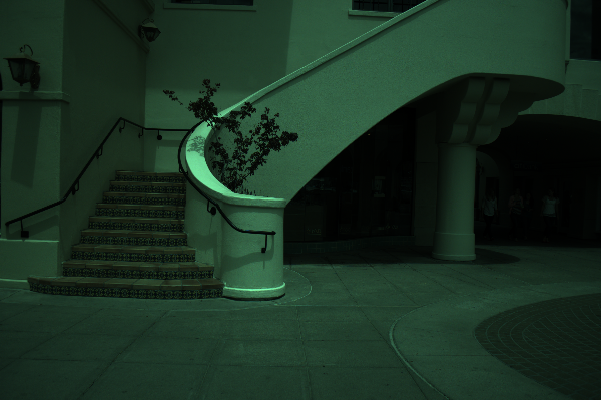} &
        \includegraphics[width=0.95\linewidth]{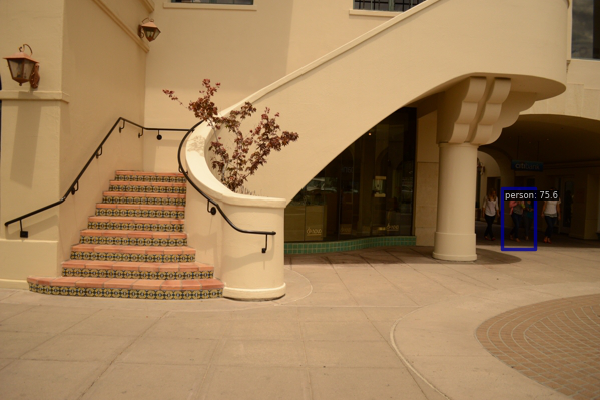} &
        \includegraphics[width=0.95\linewidth]{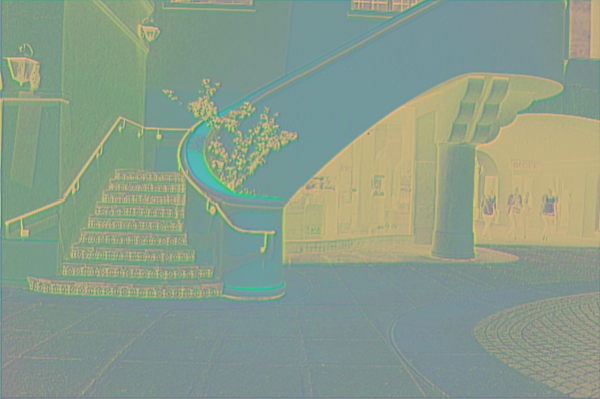} &
        \includegraphics[width=0.95\linewidth]{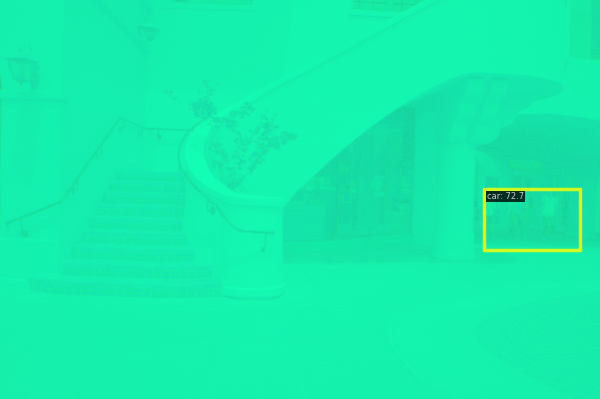} &
        \includegraphics[width=0.95\linewidth]{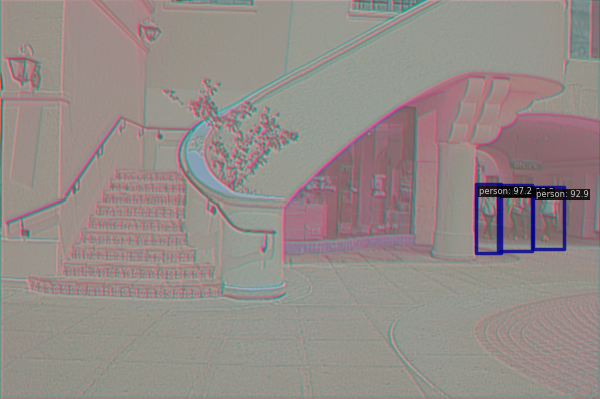} &
        \includegraphics[width=0.95\linewidth]{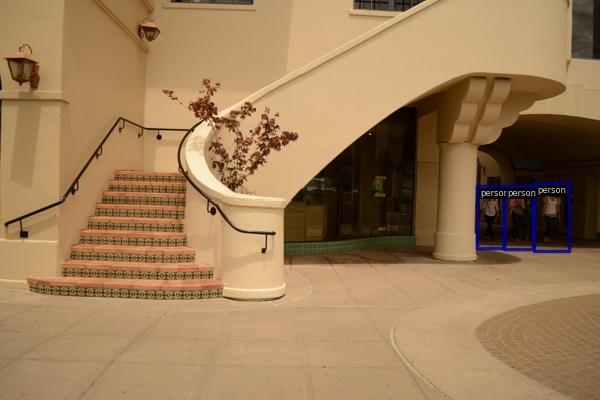}
        \\[2pt]
        
        & \multicolumn{1}{c}{{\sffamily RAW}} & \multicolumn{1}{c}{{\sffamily sRGB}} & \multicolumn{1}{c}{{\sffamily GenISP}} & \multicolumn{1}{c}{{\sffamily RAOD}} & \multicolumn{1}{c}{{\sffamily RAM (Ours)}} & \multicolumn{1}{c}{{\sffamily GT}} \\
    \end{tabular}
    \caption{Qualitative comparison of object detection results across four datasets (rows) and five approaches (columns) compared against ground truth (GT). Predictions with confidence scores above 0.5 are shown, illustrating the robustness of our approach in challenging conditions, such as occluded objects at night, transparent bottles, and small pedestrians in shadows. We display the input images after the pre-processing stage for each method. 
    }
    \label{fig:qualitative_results}
\end{figure*}

 We compare RAM against several methods that optimize ISP parameters using object detection loss, including approaches that learn and apply static or dynamic ISP parameters across datasets including Gamma \cite{10.1007/978-3-031-31435-3_25}, which applies a single parameter transformation across the entire dataset, and DynamicISP \cite{Yoshimura_2023_ICCV}, which dynamically adjusts ISP parameters. RAM outperforming them by +3.8\% and +3.1\% mAP on LOD-Normal, and +1.8\% and +4.1\% on PASCALRAW. To further highlight the advantages of RAM over existing methods, we focus on its performance on the datasets where methods like GenISP \cite{MorawskiCLDHH22} and RAOD \cite{xu2023toward} were introduced. These methods, along with IA-ISPNet \cite{liu2022image}, are the most similar to RAM, as they are also DNN-based approaches optimizing pre-processing modules on RAW data for object detection. While GenISP, optimized for the NOD dataset, achieves solid results on low-light images, RAM surpasses it by +1.2\% on NOD-Nikon and +1.7\% on NOD-Sony, demonstrating stronger handling of the challenges posed by low-light conditions. Similarly, RAOD, tailored for the ROD dataset, performs well on ROD-Night and ROD-Day, but RAM improves mAP by +3.9\% and +2.1\%, respectively, showcasing a stronger ability to manage HDR RAW data. These comparisons emphasize RAM's clear advantage in both difficult scenarios and on the specific datasets where these methods were originally developed. To illustrate these results, we present qualitative detection examples in \cref{fig:qualitative_results}.

\textbf{Evaluating RAM with Frozen Detector.}
In many cases, RAW image datasets are too small to train large object detectors effectively. A common practice is to utilize a large pre-trained detector trained on a more extensive dataset. While our method is designed for an end-to-end optimization, we acknowledge that this approach is not always feasible due to the limited size of available RAW datasets. In Table \ref{table_frozen}, we evaluate our method using a large YOLOX detector \cite{ge2021yolox} pre-trained on the COCO dataset \cite{lin2014microsoft}. Following the practice used in recent methods like AdaptiveISP \cite{wang2024adaptiveisp}, which optimize ISP parameters using RL, and GenISP \cite{morawski2022genisp}, we freeze both the backbone and the detector, training only the pre-processing stage. As the results demonstrate, RAM adapts most effectively to the pre-trained features compared to other approaches, despite the detector being originally trained on sRGB data rather than RAW.

\begin{table}
\renewcommand{\arraystretch}{1.05}
\centering
\caption{Frozen YOLOX-X with training applied only to the pre-processing stage on LOD-Dark and NOD-Sony.}
\adjustbox{max width=\linewidth}{%
\begin{tabular}{lcccccccc}
\hline
Method & \multicolumn{2}{c}{LOD-Dark} & \multicolumn{2}{c}{NOD-Sony} & \multirow{2}{*}{Params (M)} & \multirow{2}{*}{\makecell{Inference \\ Time (ms)}} \\
& mAP & mAP$_\text{50}$ & mAP & mAP$_\text{50}$ \\
\hline
sRGB & 37.9 & 52.8 & 24.4 & 42.4 & - & -\\
AdaptiveISP \cite{wang2024adaptiveisp} & 47.9 & 64.8 & 27.2 & 48.0 & 17.35 & 7.35 \\
GenISP \cite{morawski2022genisp} & 49.5 & 66.2 & 27.8 & 47.7 & 0.57 & 0.86\\
RAOD \cite{xu2023toward} & 43.1 & 58.6 & 27.4 & 47.2 & 0.28 & 1.22\\
\hline
\textbf{RAM (Ours)} & \textbf{50.8} & \textbf{66.8} & \textbf{29.6} & \textbf{51.5} & 0.54 & 1.05\\
\hline
\end{tabular}
}
\label{table_frozen}
\vspace{-2mm}
\end{table}

\subsection{Ablation Studies}
\hspace{\parindent}\textbf{Efficiency Analysis}. We show in \cref{table_flops_params} that the performance improvement of RAM cannot be attributed solely to the increase in model parameters, as demonstrated by the comparison between ResNet18 and ResNet50. While ResNet50 introduces significantly more parameters and higher latency, the performance gain is marginal compared to the boost achieved by RAM, underscoring the effectiveness of the selected architecture. Additionally, we introduce RAM-T (Tiny), an even smaller version of RAM with a reduced number of parameters while maintaining the same architecture (for more details, see supplementary material). Despite having less than half the parameters of RAM, RAM-T achieves performance remarkably close to the full module. This proximity in performance despite the smaller model size highlights that the success of the module comes from its architecture and optimization strategy, making RAM-T an excellent choice for resource-constrained applications where memory and efficiency are critical.

\textbf{Sequential vs Parallel}.
To demonstrate the advantages of our parallel pipeline over a sequential approach, we implement a sequential version that applies the same ISP functions used in RAM. Specifically, we apply separate RPE and RPD modules to each ISP function, as RPE sharing is not feasible in a sequential setup. The functions are applied in the following order—first white balance, followed by color correction, gamma correction (acting as tone mapping), and finally brightness adjustment. This final output is then passed to the backbone and detector, and the entire pipeline is trained end-to-end from scratch in an adaptive manner, similar to RAM. We compare these approaches on LOD datasets, where the parallel pipeline achieves significant mAP improvements over sequential across LOD-Dark and LOD-Normal, as shown in \cref{tab:seq_vs_par}. This is accomplished with lower inference time and fewer parameters than the sequential approach, which requires multiple RPEs.

\begin{table}
\centering
\caption{Comparison of sequential and parallel pipelines on LOD-Dark and LOD-Normal datasets.}
\adjustbox{max width=\linewidth}{%
\begin{tabular}{ccccccc}
\hline
Method & \multicolumn{2}{c}{LOD-Dark} & \multicolumn{2}{c}{LOD-Normal} & \multirow{2}{*}{Params (M)} & \multirow{2}{*}{\makecell{Inference \\ Time (ms)}}\\
& mAP & mAP$_\text{50}$ & mAP & mAP$_\text{50}$ &  &  \\
\hline
    Sequential & 29.9 & 52.0 & 35.1 & 56.8 & 1.06 & 1.38 \\
    Parallel & \textbf{35.2} & \textbf{58.4} & \textbf{40.1} & \textbf{61.4} & 0.54 & 1.05 \\
\hline
\end{tabular}
}
\label{tab:seq_vs_par}
\end{table}

\begin{figure}[t]
    \centering
    \begin{minipage}[b]{0.32\linewidth}
        \centering
        \includegraphics[width=\linewidth]{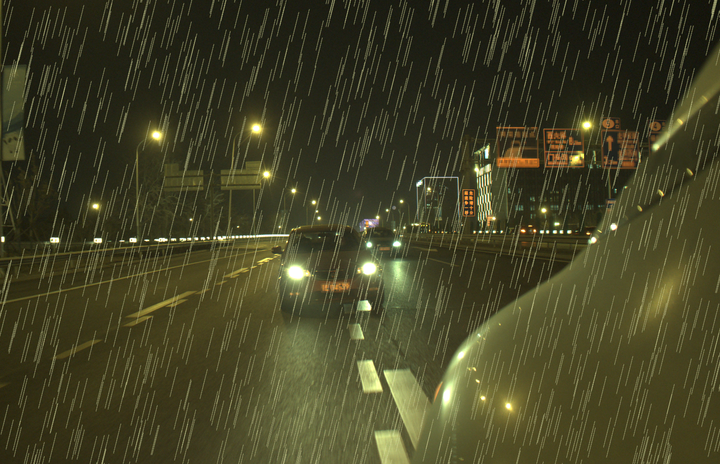}
        \subcaption{Rain}
    \end{minipage}
    \begin{minipage}[b]{0.32\linewidth}
        \centering
        \includegraphics[width=\linewidth]{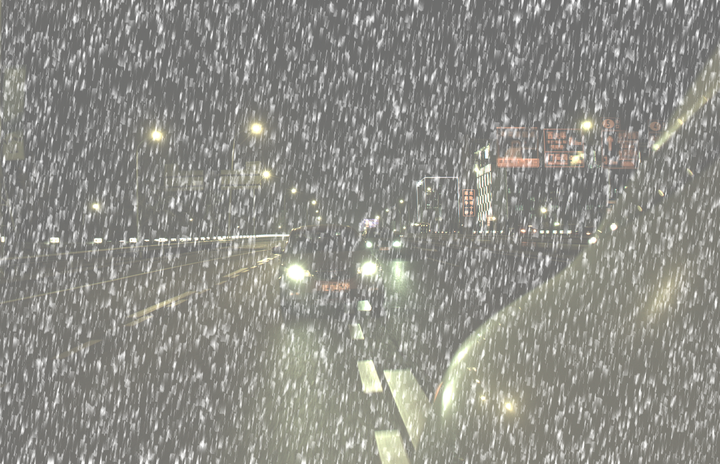}
        \subcaption{Snow}
    \end{minipage}
    \begin{minipage}[b]{0.32\linewidth}
        \centering
        \includegraphics[width=\linewidth]{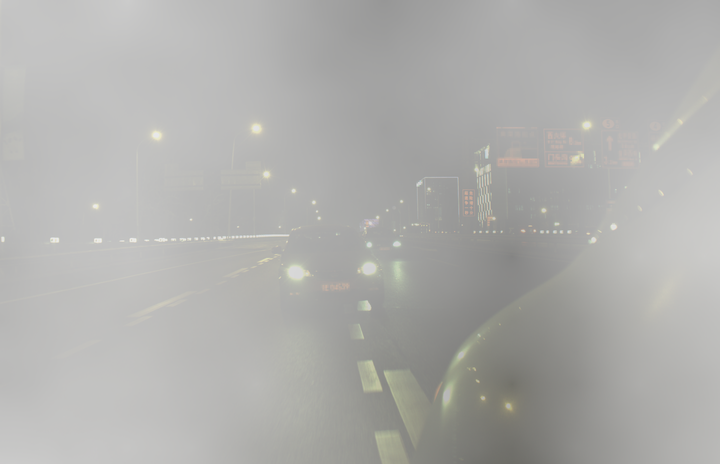}
        \subcaption{Fog}
    \end{minipage}
    
    \begin{minipage}[b]{0.32\linewidth}
        \centering
        \includegraphics[width=\linewidth]{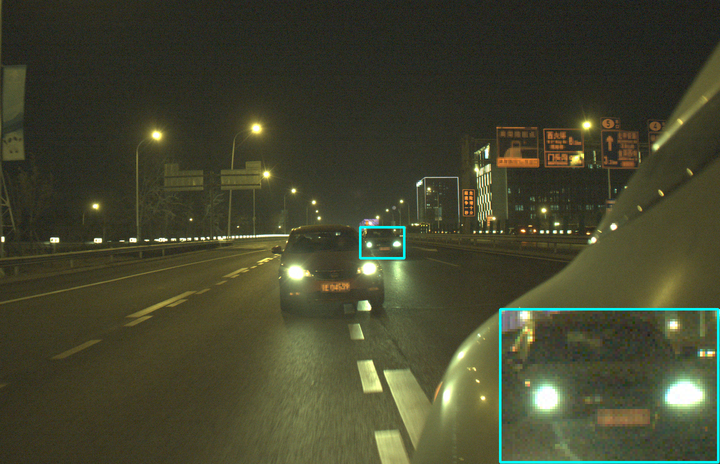}
        \subcaption{Mild}
    \end{minipage}
    \begin{minipage}[b]{0.32\linewidth}
        \centering
        \includegraphics[width=\linewidth]{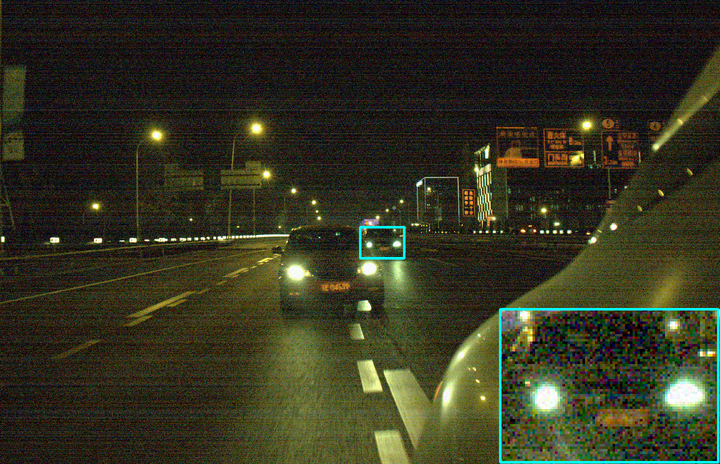}
        \subcaption{Medium}
    \end{minipage}
    \begin{minipage}[b]{0.32\linewidth}
        \centering
        \includegraphics[width=\linewidth]{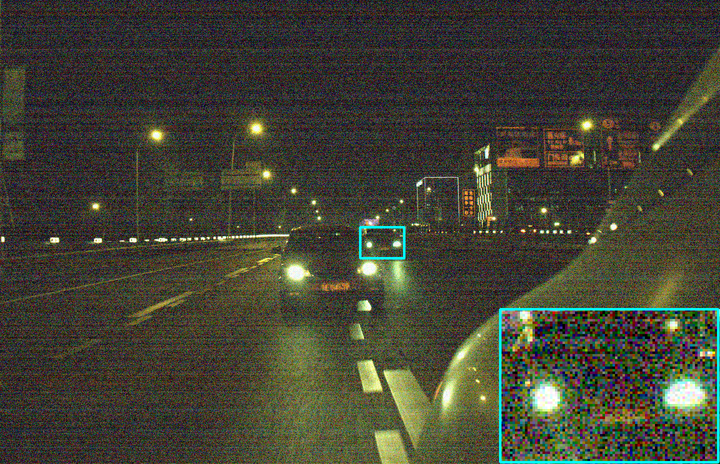}
        \subcaption{Strong}
    \end{minipage} 

    \caption{Illustration of our synthesized data on the ROD-Night dataset, demonstrating images with synthesized weather conditions in (a)-(c) and synthesized noise in (d)-(f).}
    \label{fig:syn_images}
    \squeezeup
    \vspace{-1mm}
\end{figure}

\subsection{Robust Object Detection in Challenging Conditions}
Beyond object detection in normal conditions, RAW images demonstrate significant advantages in challenging environments such as low-light, high-noise scenarios, and adverse weather. Their higher bit depth and richer information content provide a superior signal-to-noise ratio (SNR) compared to sRGB, which undergoes aggressive processing that can discard crucial details. To validate this, we simulate degraded images and evaluate detection performance, showing that while sRGB struggles with limited processed data, RAW retains crucial information. Combined with the adaptability nature of RAM, our approach ensures robust object detection in adverse environments.

\subsubsection{Low-light Object Detection in Noisy Environments}
\label{sec:noise}
We apply the state-of-the-art LED denoising algorithm~\cite{jin2023lighting} to the RAW night images from ROD~\cite{xu2023toward} for low-light noise synthesis and denoising. For each image, we synthesize three noise severity levels (Mild, Medium, and Strong) by setting the ratio parameter defined in LED to 100, 200, and 300, respectively. The resulting images are used for the noisy experiments. After synthesis, we use LED to denoise the noisy RAW images. These denoised RAW images are then given as inputs for the denoised experiments. Illustrative examples of the noisy images can be found in \cref{fig:syn_images}.

The results in \cref{tab:noise_comparison} provide a detailed comparison of object detection performance on noisy and denoised data across three different noise levels—Mild, Medium, and Strong—for RAW, sRGB, and RAM pre-processing. RAM consistently outperforms both RAW and sRGB, demonstrating its superior ability to handle noise in both noisy and denoised scenarios. In many cases, RAM even manages to achieve better detection results on noisy data than sRGB does on denoised data. This highlights how well RAM can mitigate the effects of noise without the need for explicit denoising, effectively adapting to the noisy inputs and still maximizing object detection performance.

The best results are, as expected, on denoised data, showing that reducing noise enhances detection accuracy. As noise levels increase, RAW outperforms sRGB, especially in the Medium and Strong noise settings. Traditional ISP operations often lose information or emphasize noise in noisy images, whereas RAW retains more details, allowing for better noise handling.
While RAM demonstrates enhanced performance in detecting all object sizes, the improvement is most notable in the detection of small objects, as shown by mAP$_\text{s}$. This significant performance gap illustrates the effectiveness of RAM in maintaining accuracy under challenging real-world conditions, where noise and the details of small objects can critically impact results.

\begin{table}
  \centering
    \caption{Comparison of results across RAW, sRGB, and RAM under Noisy and Denoised data for various noise levels (Mild, Medium, Strong). The results are reported using mAP and mAP$_\text{s}$ (small).}
  \adjustbox{max width=\linewidth} {
  \begin{tabular}{lccccccc}
    \toprule
    & & \multicolumn{2}{c}{Mild} & \multicolumn{2}{c}{Medium} & \multicolumn{2}{c}{Strong} \\
    \cmidrule(lr){3-4} \cmidrule(lr){5-6} \cmidrule(lr){7-8}
    Method & Data & mAP & mAP$_\text{s}$ & mAP & mAP$_\text{s}$ & mAP & mAP$_\text{s}$ \\
    \midrule
    \multirow{2}{*}{RAW} & Noisy & 44.4 & 29.6 & 42.2 & 29.0 & 40.2 & 27.4 \\
    & Denoised & 47.5 & 32.2 & 44.9 & 31.0 & \underline{43.4} & 30.2 \\
    \midrule
    \multirow{2}{*}{sRGB} & Noisy & 44.5 & 29.2 & 41.0 & 26.6 & 38.9 & 24.8 \\
    & Denoised & 47.6 & 32.7 & 44.9 & 31.0 & 42.3 & 28.7 \\
    \midrule
    \multirow{2}{*}{RAM} & Noisy & \underline{49.1} & \underline{36.1} & \underline{45.1} & \underline{32.6} & 42.4 & \underline{30.5} \\
    & Denoised & \textbf{52.0} & \textbf{39.1} & \textbf{48.5} & \textbf{35.7} & \textbf{45.7} & \textbf{33.4} \\
    \bottomrule
  \end{tabular}
  }
  \label{tab:noise_comparison}
 \vspace{-3mm}
\end{table}

\subsubsection{Handling Difficult Weather Conditions}
\label{sec:weather}

\begin{figure*}[t]
    \centering
    \begin{minipage}[b]{0.19\linewidth}
        \centering
        \includegraphics[width=0.95\linewidth]{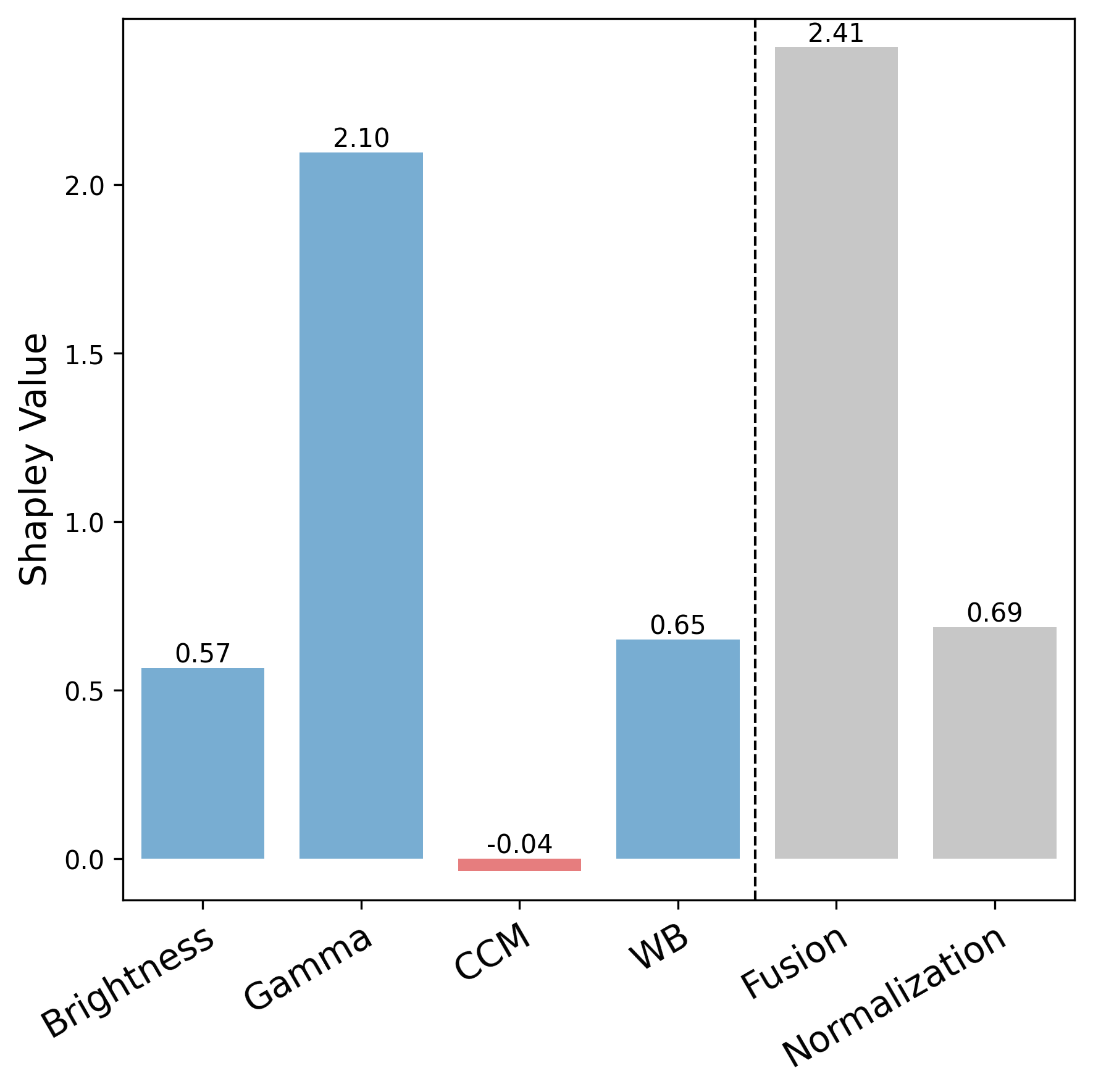}
        \subcaption{LOD-Dark}
    \end{minipage}
    \begin{minipage}[b]{0.19\linewidth}
        \centering
        \includegraphics[width=0.95\linewidth]{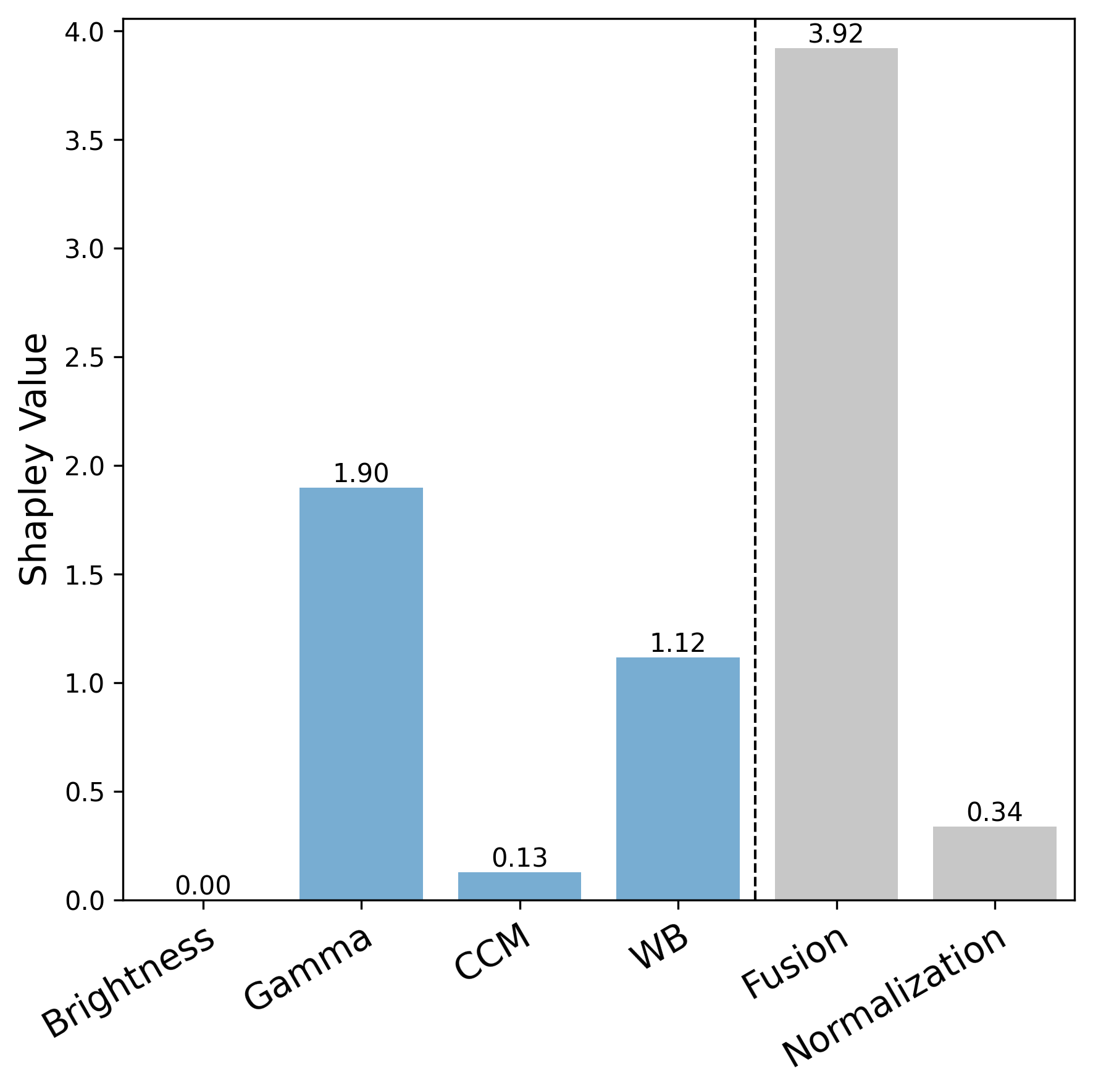}
        \subcaption{LOD-Normal}
    \end{minipage}
    \begin{minipage}[b]{0.19\linewidth}
        \centering
        \includegraphics[width=0.95\linewidth]{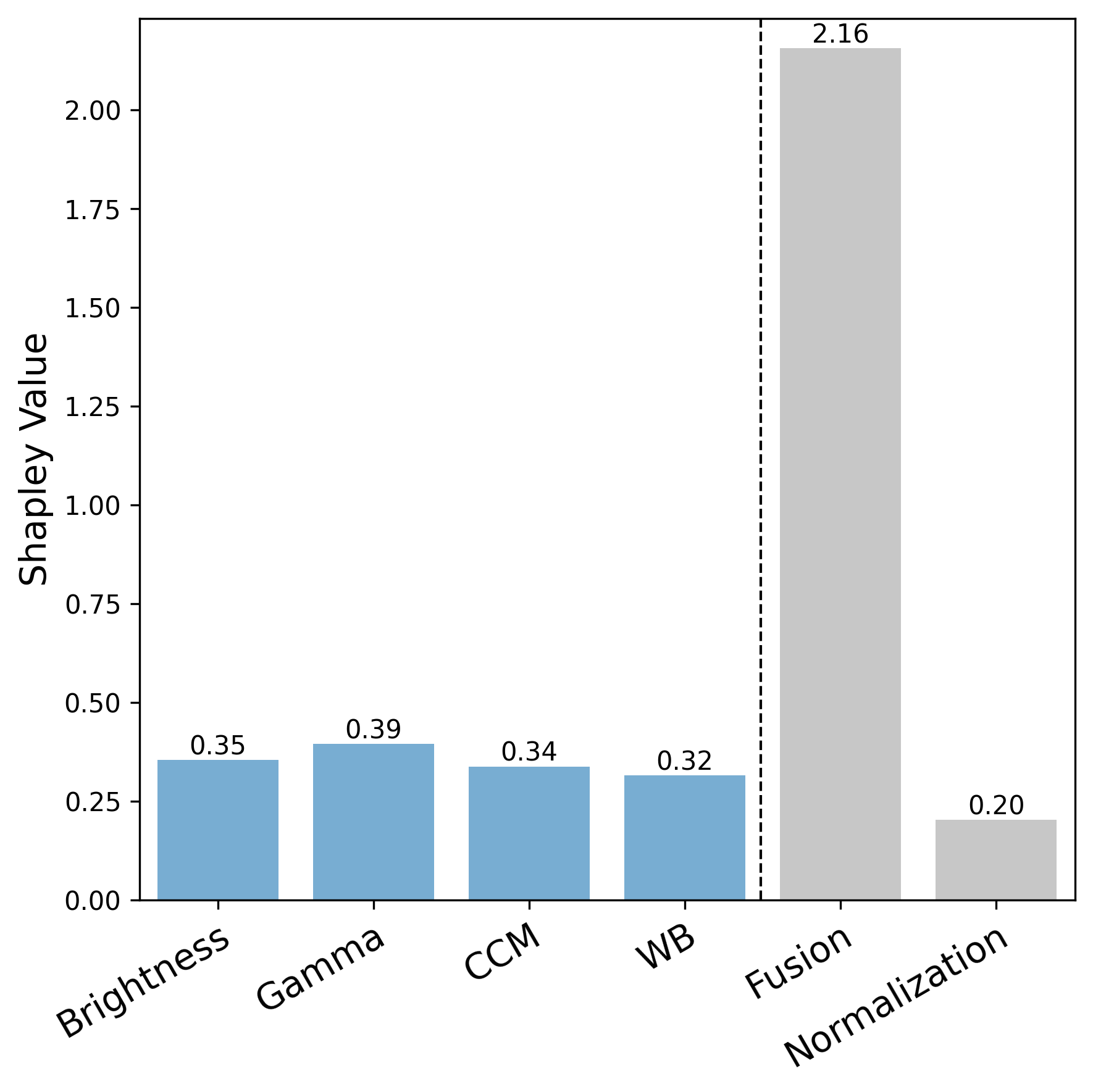}
        \subcaption{NOD-Sony}
    \end{minipage} 
    \begin{minipage}[b]{0.19\linewidth}
        \centering
        \includegraphics[width=0.95\linewidth]{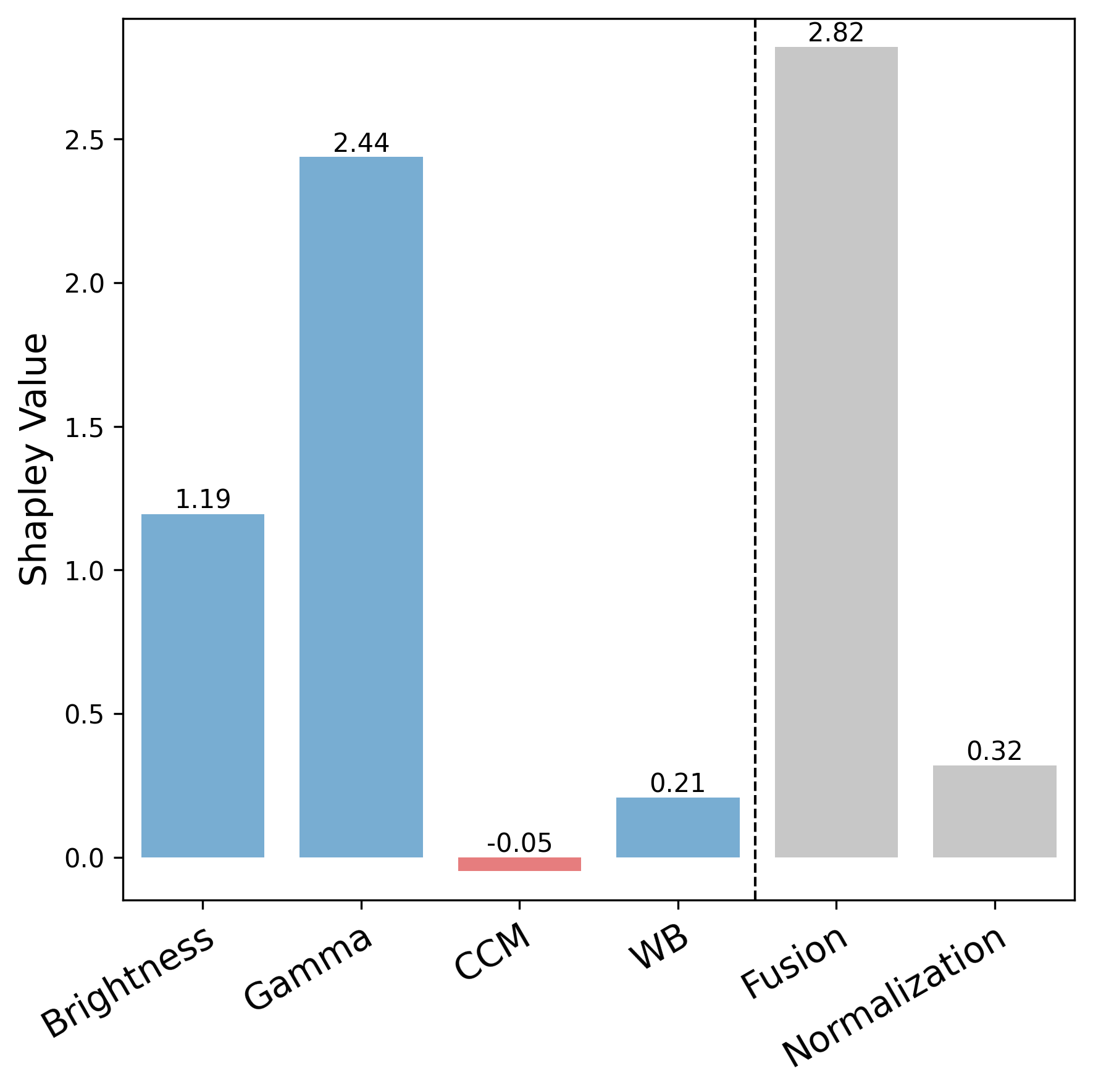}
        \subcaption{ROD-Day}
    \end{minipage}
    \begin{minipage}[b]{0.19\linewidth}
        \centering
        \includegraphics[width=0.95\linewidth]{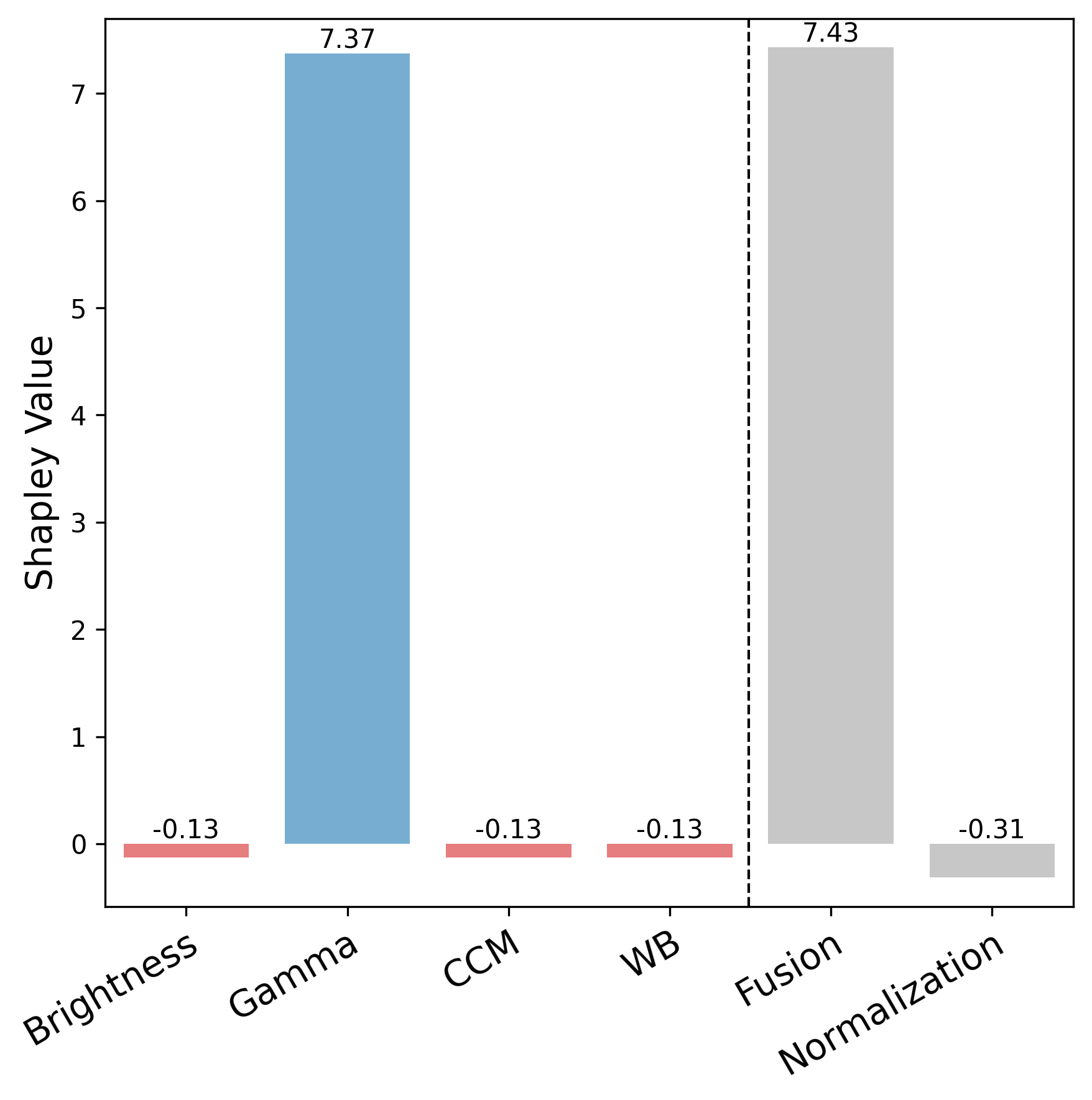}
        \subcaption{ROD-Night}
    \end{minipage} \hfill

    \caption{Shapley values analysis showing the importance of each component within the parallel RAM mechanism across different datasets. Blue bars represent ISP functions with reduced performance when a component is removed, while red bars indicate improved performance. Gray bars represent general RAM components.}
    \label{fig:shapley}
    \vspace{-2mm}
\end{figure*}

In this section, we highlight the ability of the RAM pre-processing module to outperform sRGB in handling challenging weather conditions. To test its robustness under such dynamic conditions, we employ the ROD dataset~\cite{xu2023toward} and synthesize foggy, rainy, and snowy weather conditions using the ImageCorruption Python library ~\cite{michaelis2019dragon}, with parameters adjusted for RAW images. Examples of these images are presented in \cref{fig:syn_images}.
All experiments were conducted using a DINO detector with ResNet50 as the backbone, trained from scratch on each specific weather dataset. 

As shown in \cref{tab:weather_conditions}, the results consistently demonstrate the superior performance of RAM across all weather conditions. 
While the clean dataset provides a baseline, RAW performance drops by 6.6\% to 9.9\% in adverse weather, whereas sRGB experiences an even larger decline---particularly in snow, with a decrease of up to 13.6\%.
In contrast, RAM remains remarkably stable, with only a marginal performance drop of 2.5\% to 3.5\%, underscoring its robustness. This shows that traditional ISP pipelines are highly sensitive to changes in environmental conditions, particularly when dealing with snow, where the model struggles to maintain accuracy. RAM, however, adapts its parameters based on the input, delivering consistently higher accuracy regardless of the weather. These findings highlight the need for adaptive pre-processing like RAM in dynamic conditions where traditional pipelines fail to generalize effectively.


\begin{table}[t]
\centering
\caption{Comparison of RAW, sRGB, and RAM across different synthetic weather conditions on ROD-Night dataset.}
\adjustbox{max width=\linewidth}{%
\begin{tabular}{ccccccccc}
\hline
\textbf{Data Type} & \multicolumn{2}{c}{\textbf{Clean}} & \multicolumn{2}{c}{\textbf{Rain}} & \multicolumn{2}{c}{\textbf{Snow}} & \multicolumn{2}{c}{\textbf{Fog}} \\
& mAP & mAP$_\text{50}$ & mAP & mAP$_\text{50}$ & mAP & mAP$_\text{50}$ & mAP & mAP$_\text{50}$ \\
\hline
    RAW & 49.6 & 79.3 & 43.0 & 72.8 & 39.7 & 68.1 & 42.9 & 73.3 \\
    sRGB & 51.0 & 80.8 & 45.1 & 75.4 & 37.4 & 64.8 & 41.0 & 72.2 \\
    RAM & \textbf{56.6} & \textbf{85.1} & \textbf{53.1} & \textbf{82.8} & \textbf{53.2} & \textbf{82.4} & \textbf{54.1} & \textbf{83.2} \\
\hline
\end{tabular}
}
\label{tab:weather_conditions}
\vspace{-2mm}
\end{table}

\subsection{Shapley Values Analysis}

Shapley values \cite{f99c1a45-348b-3431-979a-6234c790659b}, based on cooperative game theory, provide a measure of each player contribution by averaging the marginal impact of removing that player. In this analysis, we use Shapley values to evaluate the importance of each element within RAM, with the contribution scores derived from the mAP of the object detector.

Let:
\begin{itemize}
    \item \( \text{mAP}_{\text{base}} \) be the mAP score without RAM.
    \item \( \text{mAP}_{\text{full}} \) be the mAP score with all RAM components.
    \item \( \text{mAP}_{-i} \) be the mAP score with component \( i \) removed.
\end{itemize}
The Shapley value \( \phi_i \) for component \(i\) is:
\[
\phi_i = \left( \frac{\text{mAP}_{\text{full}} - \text{mAP}_{-i}}{\sum_{j=1}^n (\text{mAP}_{\text{full}} - \text{mAP}_{-j})} \right) \cdot (\text{mAP}_{\text{full}} - \text{mAP}_{\text{base}})
\]

The results, illustrated in \cref{fig:shapley}, show significant insights into how different components impact performance across various datasets. In both LOD datasets, gamma correction and white balance are important components. However, while brightness proves beneficial in the LOD-Dark dataset with short exposure, it has no effect in the LOD-Normal long exposure version, where sufficient exposure already supports visibility. Gamma correction is particularly effective for the ROD datasets, reflecting its role in tone mapping for HDR images. Its significance is more pronounced in ROD-Night compared to ROD-Day, where the low light levels in night images benefit more from gamma adjustment. In the case of ROD-Day, brightness adjustments help manage varying light levels in HDR scenes, improving visibility in shadowed areas. Finally, For the NOD-Sony dataset, the overall improvement is due to the combined effect of all functions along with Feature Fusion module, rather than any single component.

The Feature Fusion module emerges as the most crucial component, as evidenced by its performance impact when replaced with a single-layer convolution. This underscores its role in integrating information from other components to optimize detection. Normalization, which typically improves detection by balancing pixel intensity distributions, has a smaller impact compared to other components.

Overall, the Shapley value analysis demonstrates the varying importance of each component depending on the characteristics of each dataset, highlighting how specific functions are tailored to enhance object detection performance under different conditions.

\section{Distribution Analysis}
\begin{figure}[t]
    \centering
    
    \begin{minipage}[b]{0.02\linewidth}
        \centering
        \adjustbox{raise=4.2ex}{
            \scalebox{0.4}{\rotatebox{90}{\sffamily ROD-Day}}
        }
    \end{minipage}
    \begin{minipage}[b]{0.31\linewidth}
        \centering        \includegraphics[width=0.95\linewidth]{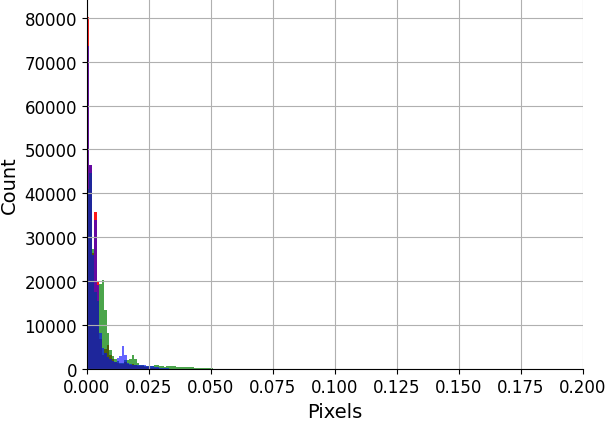}
    \end{minipage}
    \begin{minipage}[b]{0.31\linewidth}
        \centering
        \includegraphics[width=0.95\linewidth]{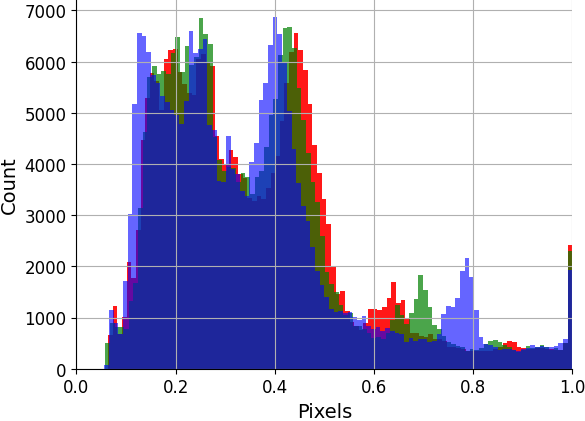}
    \end{minipage}
    \begin{minipage}[b]{0.31\linewidth}
        \centering
        \includegraphics[width=0.95\linewidth]{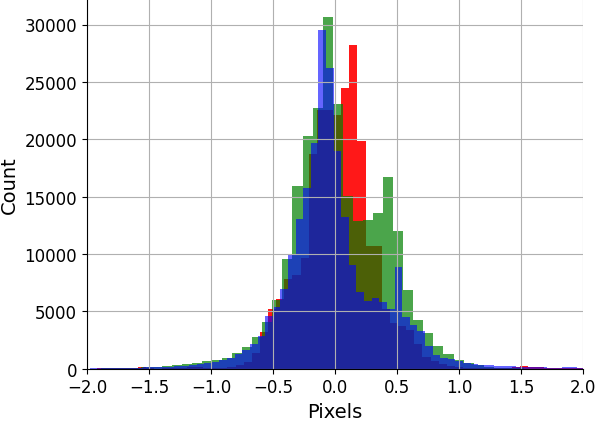}
    \end{minipage} \hfill

    \begin{minipage}[b]{0.02\linewidth}
        \centering
        \adjustbox{raise=4.0ex}{
            \scalebox{0.4}{\rotatebox{90}{\sffamily LOD-Normal}}
        }
    \end{minipage}
\begin{minipage}[b]{0.31\linewidth}
        \centering        \includegraphics[width=0.95\linewidth]{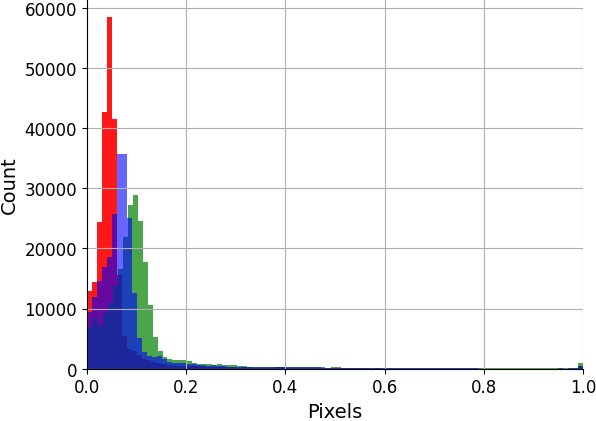}
    \end{minipage}
    \begin{minipage}[b]{0.31\linewidth}
        \centering
        \includegraphics[width=0.95\linewidth]{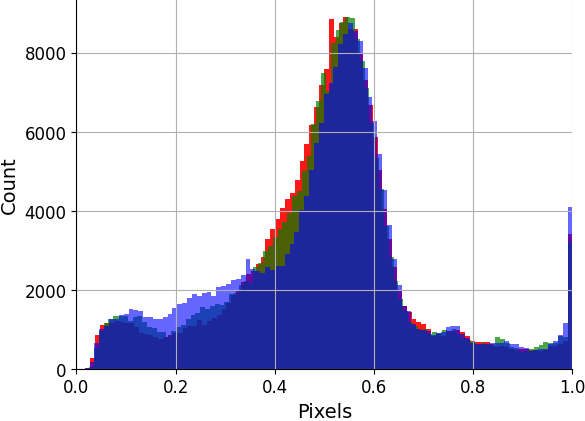}
    \end{minipage}
    \begin{minipage}[b]{0.31\linewidth}
        \centering
        \includegraphics[width=0.95\linewidth]{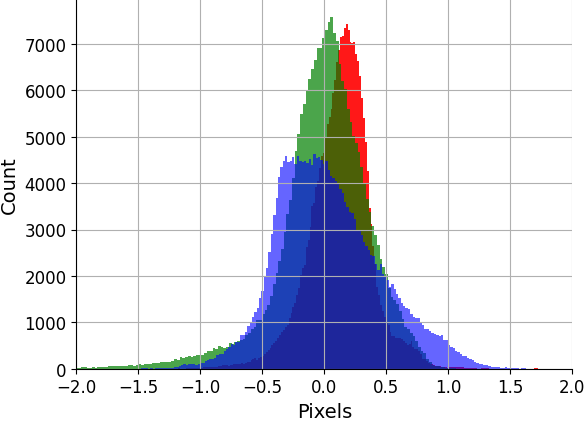}
    \end{minipage} \hfill    

        \begin{minipage}[b]{0.02\linewidth}
        \centering
        \adjustbox{raise=6.5ex}{
            \scalebox{0.4}{\rotatebox{90}{\sffamily NOD-Nikon}}
        }
    \end{minipage}
        \begin{minipage}[b]{0.31\linewidth}     
        \includegraphics[width=0.95\linewidth]{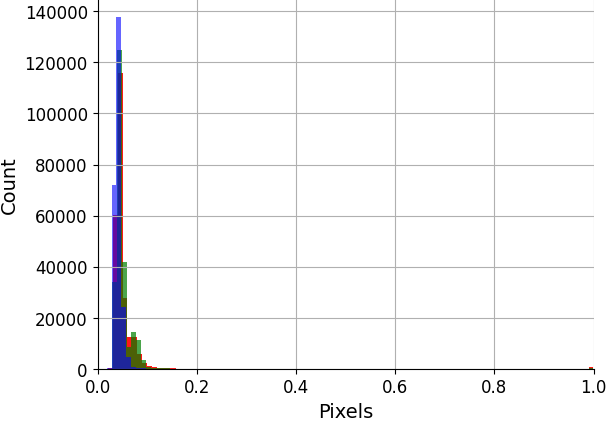}
        \subcaption{RAW}
    \end{minipage}
    \begin{minipage}[b]{0.31\linewidth}        
        \includegraphics[width=0.95\linewidth]{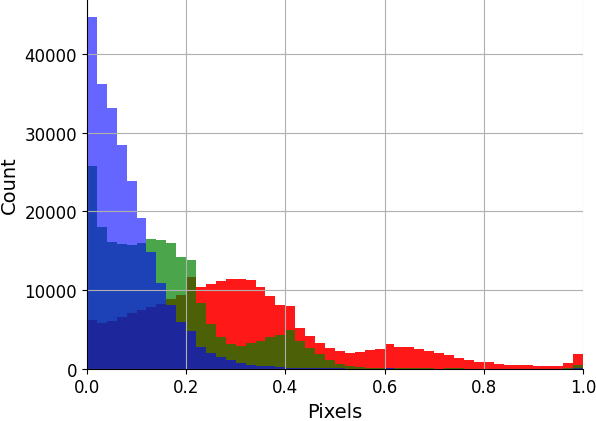}
        \subcaption{sRGB}
    \end{minipage}
    \begin{minipage}[b]{0.31\linewidth}        
        \includegraphics[width=0.95\linewidth]{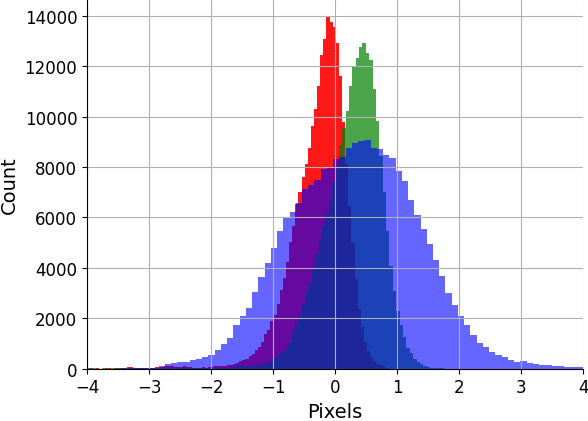}
        \subcaption{RAM (ours)}
    \end{minipage} 

    \caption{Example histograms of R, G, and B channels for RAW, sRGB, and RAM outputs. The top row shows results for ROD-Day dataset (24-bit), the middle for LOD-Normal (14-bit), and the bottom row shows results for NOD-Nikon dataset (14-bit). The RAW and sRGB data are divided by the maximum pixel value for visualization. The RAM pre-processing method produces a distribution resembling a normal distribution, centered around zero.}
    \label{fig:histograms}
\end{figure}

In \cref{fig:histograms}, we present the histograms of three data types: RAW sensor data, sRGB, and the output of our pre-processing method on multiple datasets. Observing these distributions, we note that RAW data tends to cluster near zero, making it less effective for DNN-based learning and limiting the ability of the network to extract meaningful features \cite{xu2023toward, yoshimura2024pqdynamicisp}. 
sRGB data provides a distribution that is better suited for DNN input, as the ISP processing stretches the histogram and redistributes pixel intensities across the available range by enhancing the darker pixels and increasing global contrast.
However, this processing often leads to information loss and reduced dynamic range. An example of this information loss can be seen in the histograms (column b of \cref{fig:histograms}), where pixel values cluster near saturation, leading to clipping.
On the other hand, we observed that the pixel distribution produced by RAM resembles a normal distribution, centered around zero—a behavior that has been discussed and shown experimentally \cite{ioffe2015batch, lei2016layer, salimans2016weight, rezende2015variational} to better support the ability of the network to learn and converge efficiently during training. This behavior arises as a byproduct of the complex, learned transformation performed by RAM, which go beyond simple normalization methods, such as scaling and shifting.

\section{Conclusions}
In this work, we introduced the Raw Adaptation Module (RAM), a novel RAW image pre-processing module for object detection. Inspired by the parallel processing mechanism of the human visual system, our model architecture employs parallel ISP functions to dynamically select only the necessary features to enhance detection performance while maintaining efficiency.

Through extensive evaluations, we demonstrated that RAM consistently outperforms state-of-the-art methods by a wide margin. Our findings emphasize that RAW data, when processed with our approach, offers significant advantages over sRGB images, especially for edge devices and applications with access to full sensor data. While our study focused on object detection, we believe RAM holds substantial potential to improve a wide range of vision tasks, surpassing conventional sRGB data in various applications. Its efficiency and compatibility with different architectures make it a practical choice for both research and deployment, particularly in resource-limited environments.

{
    \small
    \bibliographystyle{ieeenat_fullname}
    \bibliography{main}
}

\clearpage
\setcounter{page}{1}
\maketitlesupplementary

\setcounter{section}{0} 
\renewcommand\thesection{\Alph{section}} 

\section{Experimental Setup}

We use the following settings for training and evaluation in our experiments presented in Section 4: Faster R-CNN is trained with SGD \cite{ruder2016overview} (lr=0.02, momentum=0.9, weight decay=$1e-4$), using a multi-step learning rate scheduler with decay at epochs 29, 49, and 79, for a total of 100 epochs. DINO is trained with AdamW \cite{Loshchilov2017DecoupledWD} (lr=$2e-4$, weight decay=$1e-4$) and a multi-step learning rate scheduler with decay at epochs 27 and 33, for 50 epochs. Data augmentation included random resizing and horizontal flipping. For RAM, inputs are downsampled to $256\times256$ before the RPE module, and RPD params outputs are applied to the original input images. We use mean-std normalization, calculated on the dataset, for both RAW and sRGB experiments. Performance is evaluated using mean average precision (mAP) across thresholds 0.5:0.95 and 0.5.

\section{Additional Evaluations}

\begin{table}
\centering
\caption{Experimental results comparing RAM applied to RAW images vs RAM applied to sRGB images.}
\adjustbox{max width=\linewidth}{%
\begin{tabular}{ccccc}
\hline
Method & \multicolumn{2}{c}{LOD-Dark} & \multicolumn{2}{c}{LOD-Normal} \\
& mAP & mAP$_\text{50}$ & mAP & mAP$_\text{50}$  \\
\hline
    RAW & 28.5 & 50.8 & 32.7 & 53.8\\
    sRGB & 28.7 & 51.2 & 34.5 & 57.0 \\
    \hline
    RAM (sRGB) & 32.9 & 56.0 & 37.9 & 60.4 \\
    RAM (RAW) & \textbf{34.9} & \textbf{57.6} & \textbf{40.1} & \textbf{61.4} \\
\hline
\end{tabular}
}
\label{tab:raw_vs_rgb}
\end{table}

\subsection{RAM: RAW vs RGB}
RAM is designed as a RAW-specific pre-processing module that optimizes ISP functions for object detection by leveraging the rich, unprocessed information available in RAW images. However, one might question whether RAM’s effectiveness is inherently tied to RAW data or if similar improvements could be achieved by applying it to standard RGB images. 
We believe the strength of our approach lies in its ability to learn and adapt ISP operations directly from unprocessed sensor data. While sRGB images have already undergone fixed ISP processing, RAW data preserves the complete sensor information, allowing RAM to discover optimal processing parameters specifically for detection tasks.

To validate this intuition empirically, we compare the performance of RAM when applied to RAW images versus sRGB images. The results on \cref{tab:raw_vs_rgb} show that while applying RAM to sRGB can improve the image for detection, it does not perform as well as RAM on RAW. This emphasizes that RAM is specifically tailored to utilize the full potential of RAW data, making it an effective pre-processing solution for RAW-based object detection.

\subsection{Scalability Across Different Detector Sizes}

\begin{table}
  \centering
    \caption{Comparison of sRGB and RAM performance across different YOLOX model sizes (Small, Medium, and Large) on the ROD-Night dataset.}
  \renewcommand{\arraystretch}{0.85} 
  \setlength{\tabcolsep}{5pt} 
  \begin{tabular}{lccc}
    \toprule
    Method & Data Type & mAP & mAP$_\text{50}$ \\
    \midrule
    \multirow{3}{*}{YOLOX-S}
    & RAW & 46.7 & 72.3  \\
    & sRGB & 50.3 & 76.4  \\
    & \textbf{RAM} & \textbf{54.8} & \textbf{80.7} \\
    \midrule
    \multirow{3}{*}{YOLOX-M}
    & RAW & 50.9 & 76.4 \\
    & sRGB & 54.8 & 79.7 \\
    & \textbf{RAM} & \textbf{58.0} & \textbf{83.0} \\
    \midrule
    \multirow{3}{*}{YOLOX-L} 
    & RAW & 53.3 & 78.0 \\
    & sRGB & 56.9 & 81.9  \\
    & \textbf{RAM} & \textbf{59.8} & \textbf{84.0}  \\
    \bottomrule
  \end{tabular}
  \label{tab:yolox}
\end{table}

To test the scalability of our approach, we evaluate its performance on different YOLOX detector sizes (Small, Medium, and Large) trained from scratch on the ROD-Night dataset. Each model size corresponds to a different number of parameters, allowing us to examine how RAM adapts across varying computational capacities.

The results in \cref{tab:yolox} show that RAM consistently outperforms sRGB across all model sizes, demonstrating its adaptability and effectiveness regardless of the detector’s scale. Notably, the performance gains achieved by integrating RAM surpass those obtained by simply increasing the detector size when using sRGB. For instance, YOLOX-S with RAM achieves higher mAP$_\text{50}$ than YOLOX-M with sRGB, and similarly, YOLOX-M with RAM outperforms YOLOX-L with sRGB. This highlights the significant impact of optimizing RAW pre-processing, confirming that our efficient approach improves detection performance more effectively than just increasing the model size.

\begin{table}
\centering
  \renewcommand{\arraystretch}{1.15} 
\caption{Comparison of training each dataset separately vs training one model on all datasets. Results are reported using mAP.}
\adjustbox{max width=\linewidth}{%
\begin{tabular}{lcccc}
\hline
Training Approach & PASCALRAW & NOD-Nikon & NOD-Sony & LOD-Dark \\
\hline
    Single dataset & 67.7 & 31.9 & 32.8 & 48.0 \\
    All datasets & \textbf{69.6} & \textbf{32.8} & \textbf{35.8} & \textbf{48.2} \\
\hline
\end{tabular}
}
\label{tab:generalization}
\end{table}

\begin{table*}
\centering
\caption{Comparison between RAM and RAM-T across different RAW object detection datasets. Results are reported using mean Average Precision (mAP) and mAP at 50\% IoU (mAP50).}
\adjustbox{max width=\linewidth}{%
\begin{tabular}{lcccccccccccccc}
\hline
\multirow{2}{*}{Method} & \multicolumn{2}{c}{ROD-Day} & \multicolumn{2}{c}{ROD-Night} & \multicolumn{2}{c}{NOD-Nikon} & \multicolumn{2}{c}{NOD-Sony} & \multicolumn{2}{c}{LOD-Dark} & \multicolumn{2}{c}{LOD-Normal} & \multicolumn{2}{c}{PASCALRAW} \\
& mAP & mAP$_\text{50}$ & mAP & mAP$_\text{50}$ & mAP & mAP$_\text{50}$ & mAP & mAP$_\text{50}$ & mAP & mAP$_\text{50}$ & mAP & mAP$_\text{50}$ & mAP & mAP$_\text{50}$ \\
\hline
RAM & \textbf{28.3} & \textbf{45.1} & \textbf{44.5} & \textbf{69.0} & \textbf{31.0} & \textbf{56.3} & \textbf{32.4} & \textbf{59.1} & 34.9 & 57.6 & \textbf{40.1} & \textbf{61.6} & \textbf{66.4} & \textbf{92.3} \\
RAM-T & 27.9 & 44.4 & 44.2 & 68.5 & 30.7 & 55.8 & 32.2 & 58.1 & \textbf{35.8} & \textbf{58.4} & \textbf{40.1} & 61.4 & 66.3 & 92.2 \\
\hline
\end{tabular}
}
\label{ram_efficient_table}
\end{table*}

\begin{table*}[]
\centering
\caption{Layer configurations for RAM and RAM-T. ConvBlock includes: Conv2d, BatchNorm2d and LeakyReLU layers.}
\begin{tabular}{|c|c|c|c|c|}
\hline
Module & Layer & Type & RAM & RAM-T \\
\hline
\multirow{3}{*}{RPEncoder} 
 & 1 & ConvBlock & 3$\rightarrow$16 channels, 7x7 kernel & 3$\rightarrow$16 channels, 3x3 kernel \\
 & 2 & MaxPool & 2x2 kernel & 2x2 kernel \\
 & 3 & ConvBlock & 16$\rightarrow$32 channels, 5x5 kernel & 16$\rightarrow$32 channels, 3x3 kernel \\
 & 4 & MaxPool & 2x2 kernel & 2x2 kernel \\
 & 5 & ConvBlock & 32$\rightarrow$128 channels, 3x3 kernel & 32$\rightarrow$64 channels, 3x3 kernel \\
 & 6 & MaxPool & 2x2 kernel & 2x2 kernel \\
 & 7 & AdaptiveAvgPool2d & 1x1 kernel & 1x1 kernel \\
\hline
\multirow{2}{*}{RPDecoder} 
 & 1 & Linear & 128 units & 64 units \\
 & 2 & LeakyReLU & - & - \\
 & 3 & Linear & 128 units & 64 units \\
\hline
\multirow{2}{*}{Feature Fusion} 
 & 1 & ConvBlock & 12$\rightarrow$16 channels, 3x3 kernel & 12$\rightarrow$16 channels, 3x3 kernel \\
 & 2 & ConvBlock & 16$\rightarrow$64 channels, 3x3 kernel & 16$\rightarrow$32 channels, 3x3 kernel \\
 & 3 & ConvBlock & 64$\rightarrow$16 channels, 3x3 kernel & 32$\rightarrow$16 channels, 3x3 kernel \\
 & 4 & Conv2D & 16$\rightarrow$3 channels, 1x1 kernel & 16$\rightarrow$3 channels, 1x1 kernel \\
\hline
\end{tabular}
\label{tab:model_modules}
\end{table*}

\subsection{Generalization Across Datasets}
\label{generalized_ram}
To demonstrate RAM’s ability to generalize across diverse data, we compare training a separate model on each dataset with training a single model jointly on all four datasets, which vary significantly in dynamic range and lighting conditions. This setting is especially challenging for RAW data: unlike RGB datasets that typically share a standardized 8-bit dynamic range, RAW datasets differ widely in their dynamic ranges and sensor-specific properties. Nevertheless, as shown in \cref{tab:generalization}, RAM adapts effectively by generating optimal parameters for each input image, enabling it to handle such variability. It not only performs well across individual datasets, but also benefits from the additional data, despite distribution differences, achieving improved performances overall.

\section{RAM vs RAM-T}

\subsection{Model Configuration}
\cref{tab:model_modules} details the layer configurations for both RAM and RAM-T (Tiny) architectures, with primary differences lying in the number of channels and kernel sizes, which ultimately affect the overall FLOPs and parameter count. The RPEncoder performs the main processing by transforming the input image into a compact feature vector, which is then fed into each RPDecoder. The RPDecoder is lightweight, making it efficient to add additional ISP functions to the pipeline without a significant computational cost. The Feature Fusion module employs a reverse-hourglass design, where the input and output channels are the smallest, while the middle layers are the largest. This structure enables efficient fusion of all processed inputs, capturing the most essential features needed for high-quality object detection.

\subsection{Quantitative Evaluation}
The comparison between RAM and RAM-T across multiple RAW object detection datasets shown in \cref{ram_efficient_table} demonstrates the effectiveness of RAM-T in achieving near-identical performance to the full RAM module. While RAM shows superior performance on most datasets, RAM-T remains a highly competitive alternative, providing a nearly equivalent detection quality with fewer parameters and reduced memory usage as shown in section 4.4. This makes RAM-T especially suitable for scenarios where computational resources and memory are more constrained, without a significant sacrifice in accuracy.

\section{Additional Visualizations}
In this section, we present additional visualizations comparing the detection results of models trained on RAW, sRGB and our method, RAM. The images in these visualizations represent different conditions synthesized on the ROD-Night dataset.

\begin{figure*}
    \centering
    \tiny
    \setlength{\tabcolsep}{0pt} 
    \begin{tabular}{m{0.03\linewidth}*{4}{>{\centering\arraybackslash}m{0.24\linewidth}}} 

        \rotatebox{90}{\sffamily RAIN} &
        \includegraphics[width=0.99\linewidth]{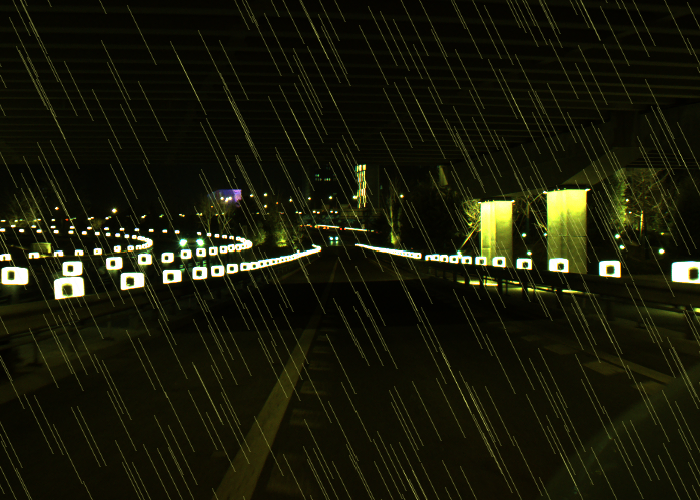} &
        \includegraphics[width=0.99\linewidth]{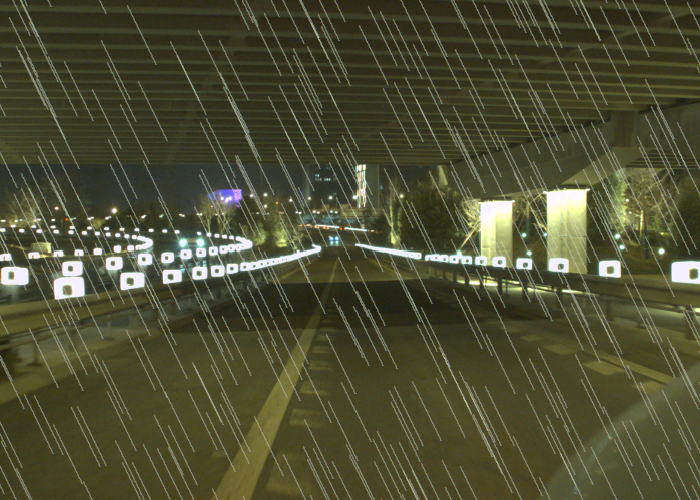} &
        \includegraphics[width=0.99\linewidth]{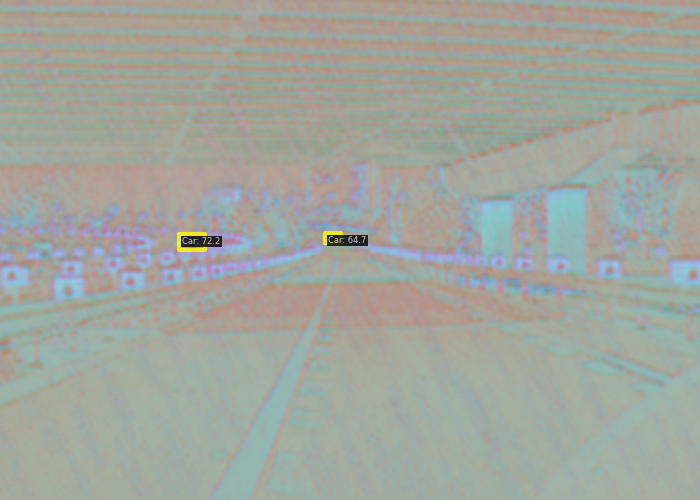} &
        \includegraphics[width=0.99\linewidth]{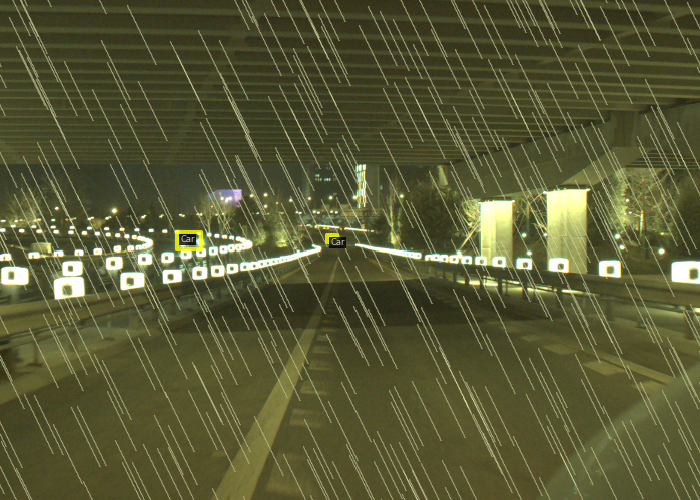} 
        \\[2pt]
        
        \rotatebox{90}{\sffamily SNOW} &
        \includegraphics[width=0.99\linewidth]{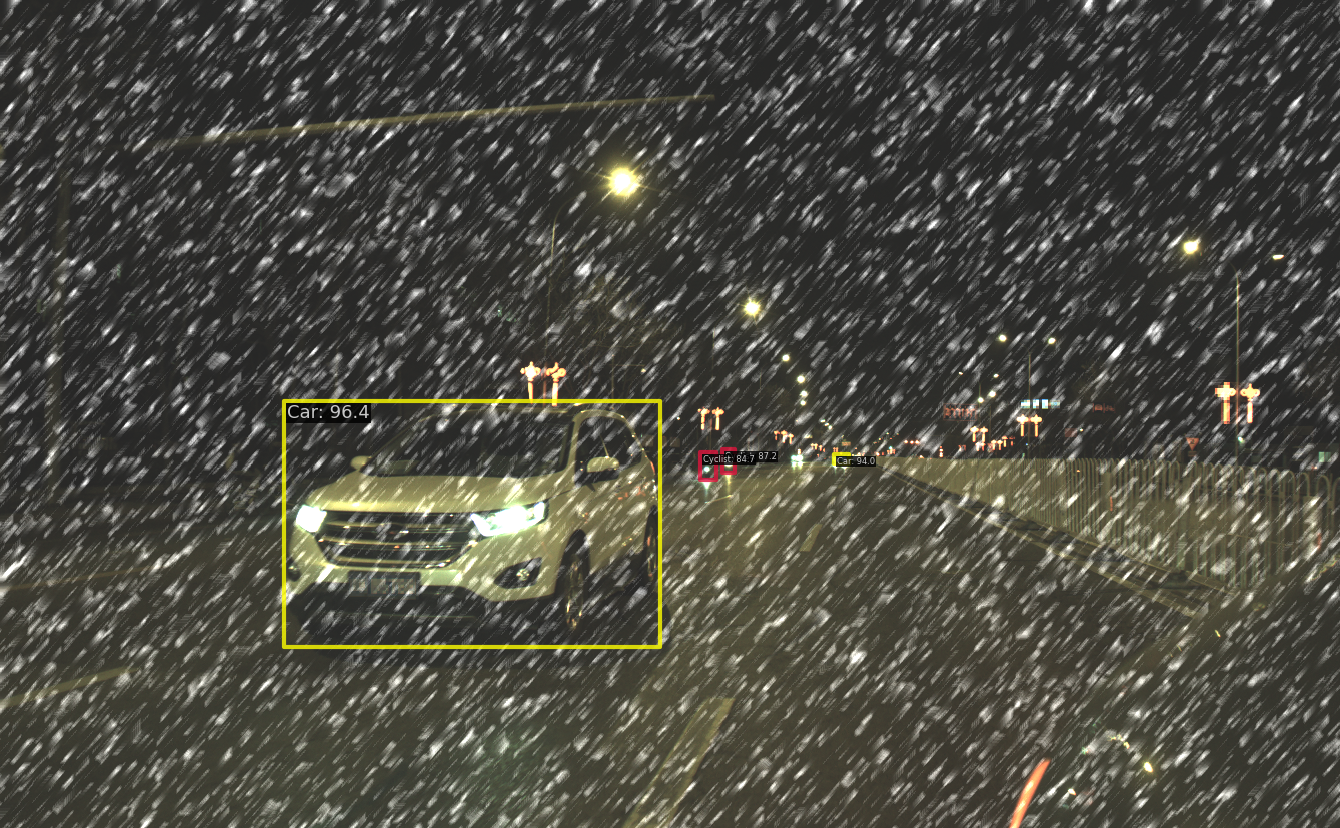} &
        \includegraphics[width=0.99\linewidth]{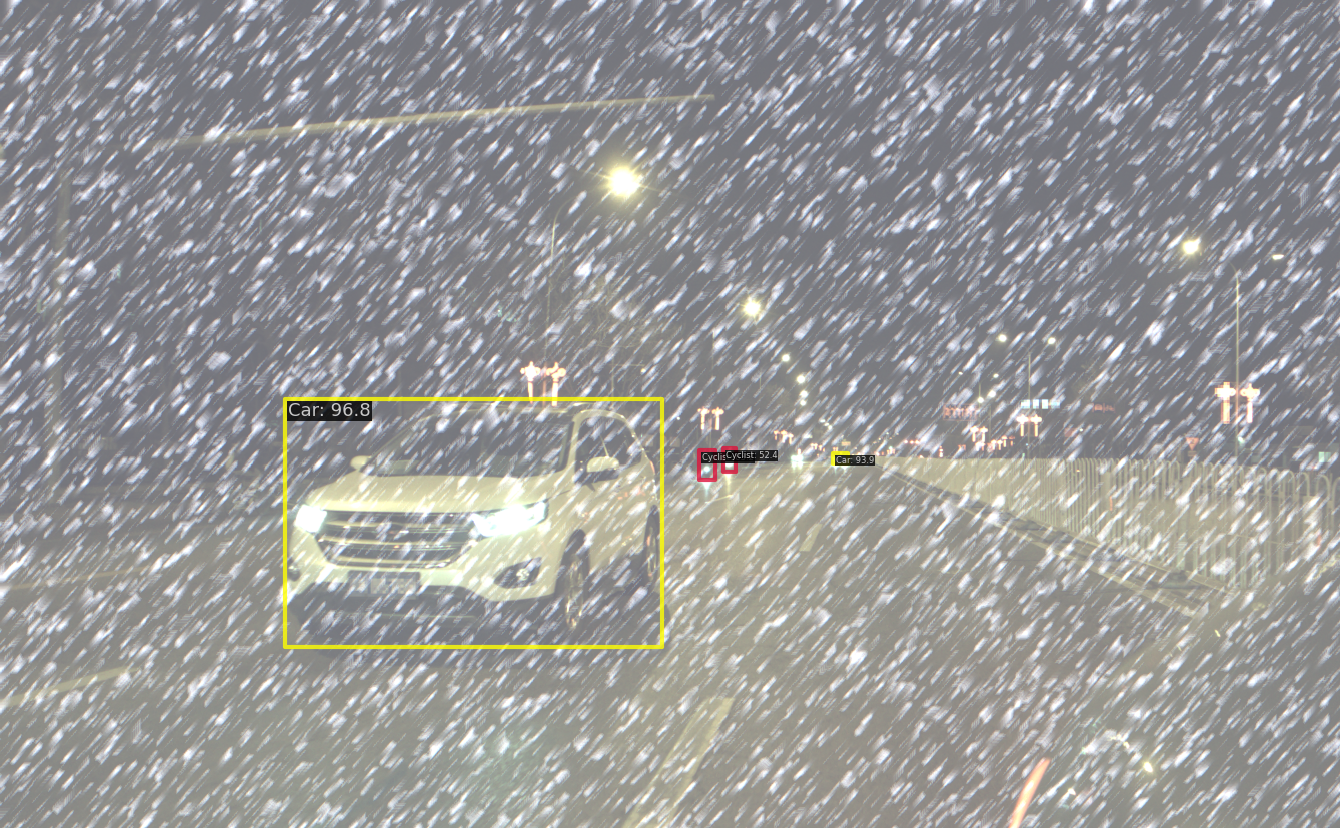} &
        \includegraphics[width=0.99\linewidth]{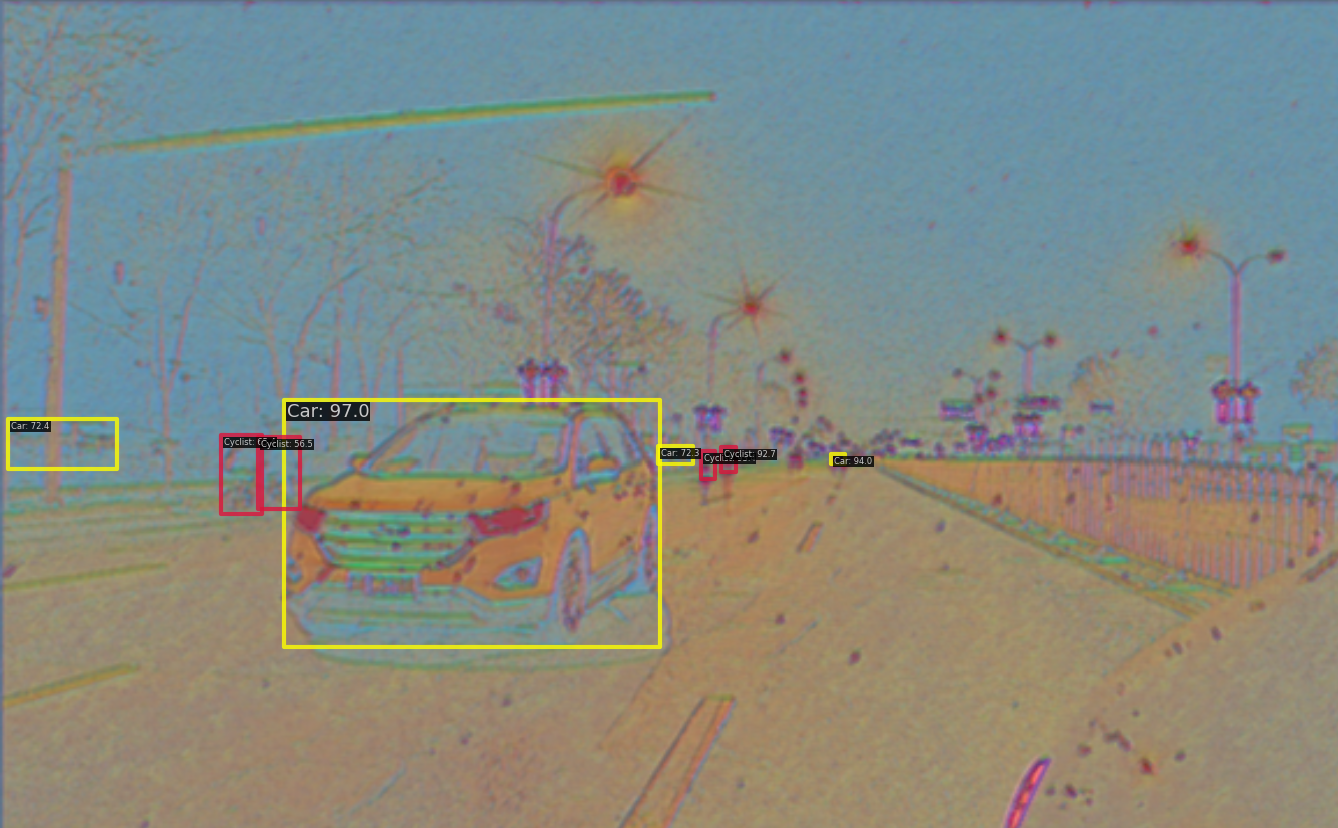} &
        \includegraphics[width=0.99\linewidth]{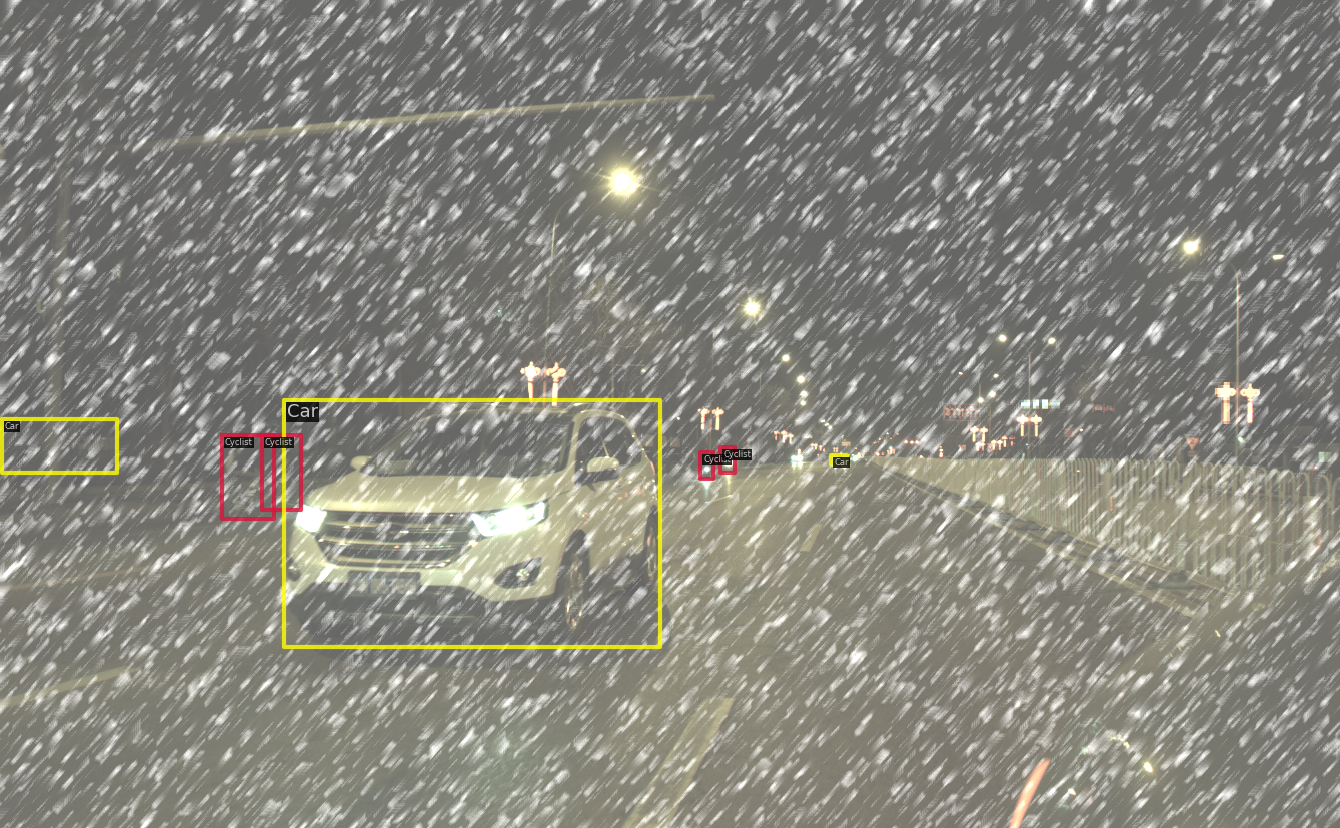}
        \\[2pt]
        
        \rotatebox{90}{\sffamily FOG} &
        \includegraphics[width=0.99\linewidth]{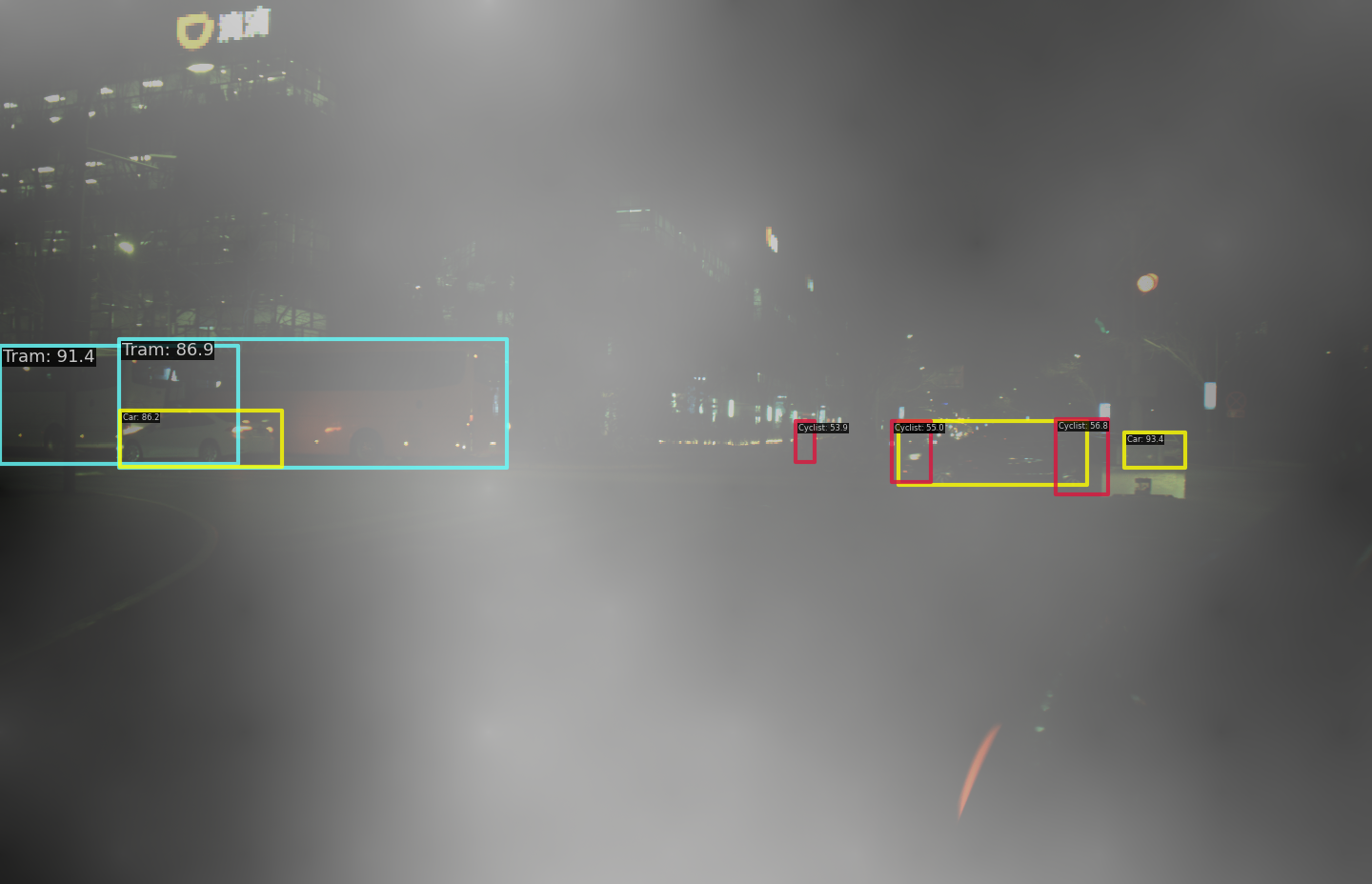} &
        \includegraphics[width=0.99\linewidth]{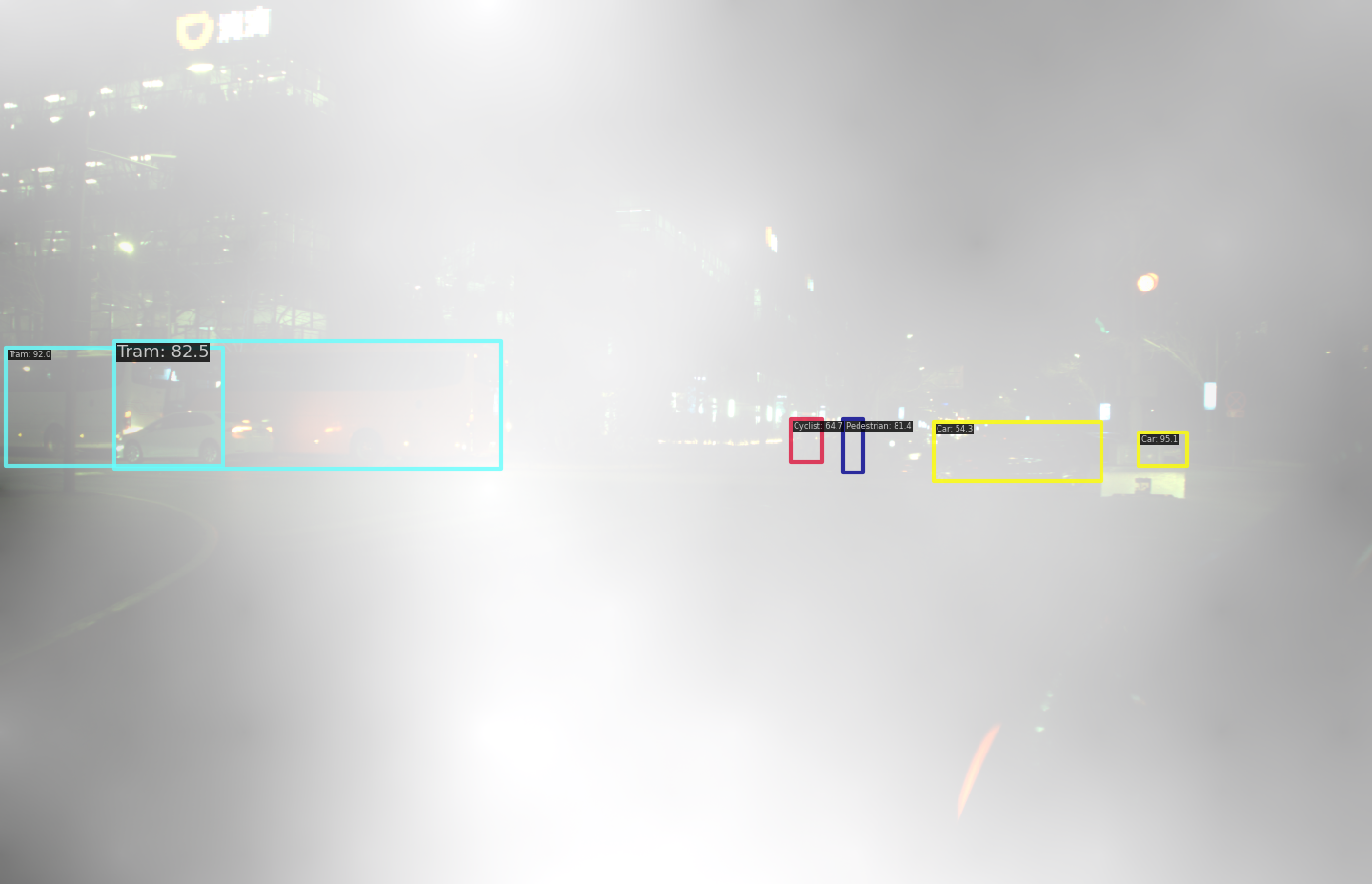} &
        \includegraphics[width=0.99\linewidth]{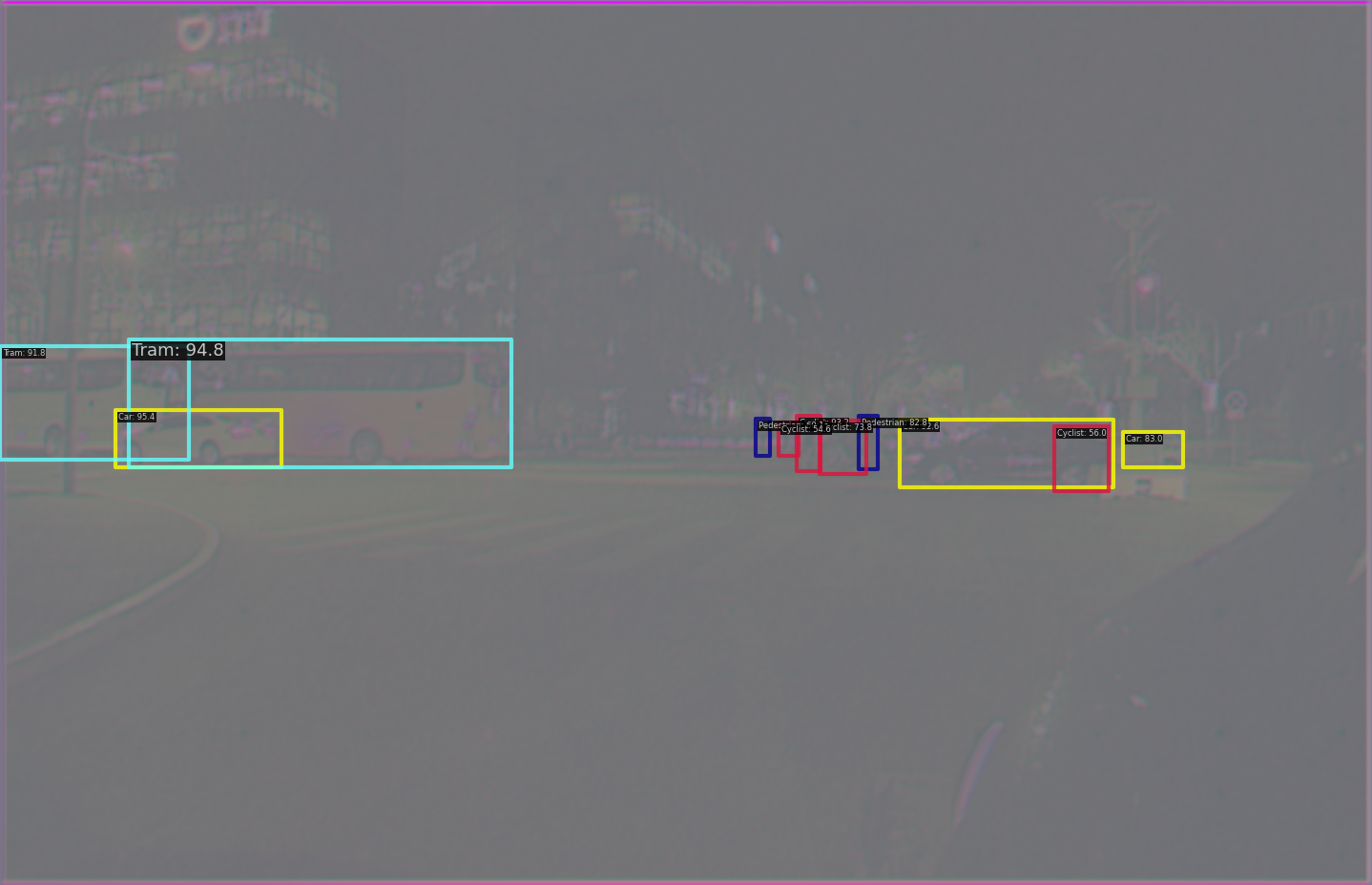} &
        \includegraphics[width=0.99\linewidth]{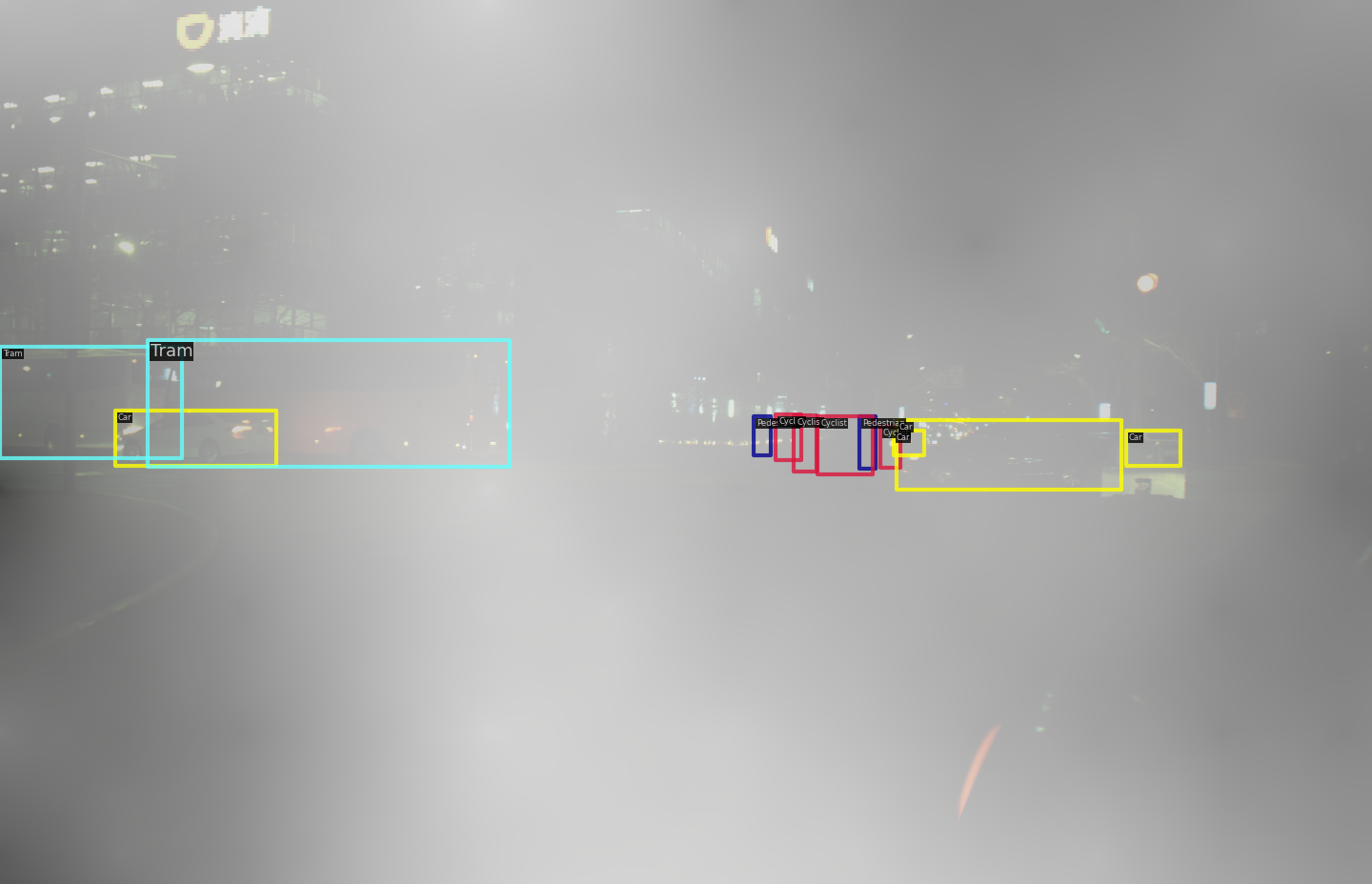}
        \\[2pt]
        
         & \multicolumn{1}{c}{{\sffamily RAW}} & \multicolumn{1}{c}{{\sffamily sRGB}} 
         & \multicolumn{1}{c}{{\sffamily RAM (ours)}}
         & \multicolumn{1}{c}{{\sffamily GT}}
         \\
    \end{tabular}
    \caption{Object detection results in challenging weather conditions—rain, snow, and fog—synthesized on the ROD-Night dataset. The columns, in left-to-right order, show RAW, sRGB, RAM (our method), and the ground truth (GT).}
    \label{fig:weather_fig}
\end{figure*}

The visualizations, shown in \cref{fig:weather_fig}, illustrate the effects of synthesized weather conditions—rain, snow, and fog—on object detection performance, corresponding to the experiments discussed in section 4.5.2.
The images in the first row demonstrate the impact of rain on detection, where the RAW and sRGB models struggle to detect small cars in rainy conditions, while RAM successfully identifies these objects. In the second row, heavy snow in the night images hides the cyclists on the left, making them nearly undetectable even to the human eye. Remarkably, RAM’s output shows an ability to “remove” most of the snow from the image, despite not being explicitly trained to do so, allowing it to accurately capture the cyclists features. In the third row, fog covers the entire scene unevenly, making detection challenging for the model. Although RAW and sRGB manage to detect some objects, their performance falls short of RAM, which generates a more consistent image less impacted by the fog.

\begin{figure*}[]
    \centering
    \tiny
    \setlength{\tabcolsep}{-1pt} 
    \begin{tabular}{c}
        \multicolumn{1}{c}{\sffamily GT} \\
        \includegraphics[width=0.4\linewidth]{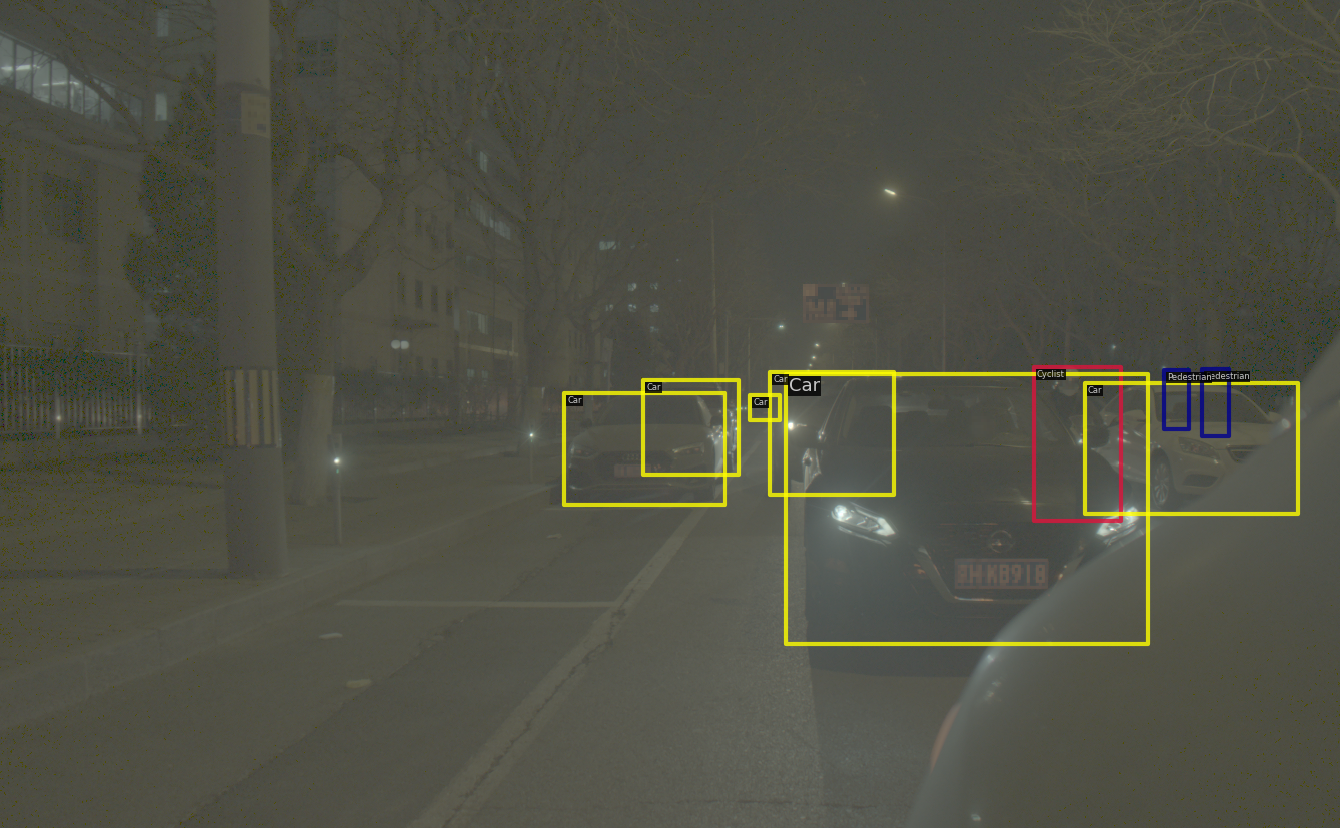} \\[2pt]
    \end{tabular}

    \begin{tabular}{m{0.03\linewidth}*{3}{>{\centering\arraybackslash}m{0.3\linewidth}}}

        \rotatebox{90}{\sffamily Noisy} &
        \includegraphics[width=0.98\linewidth]{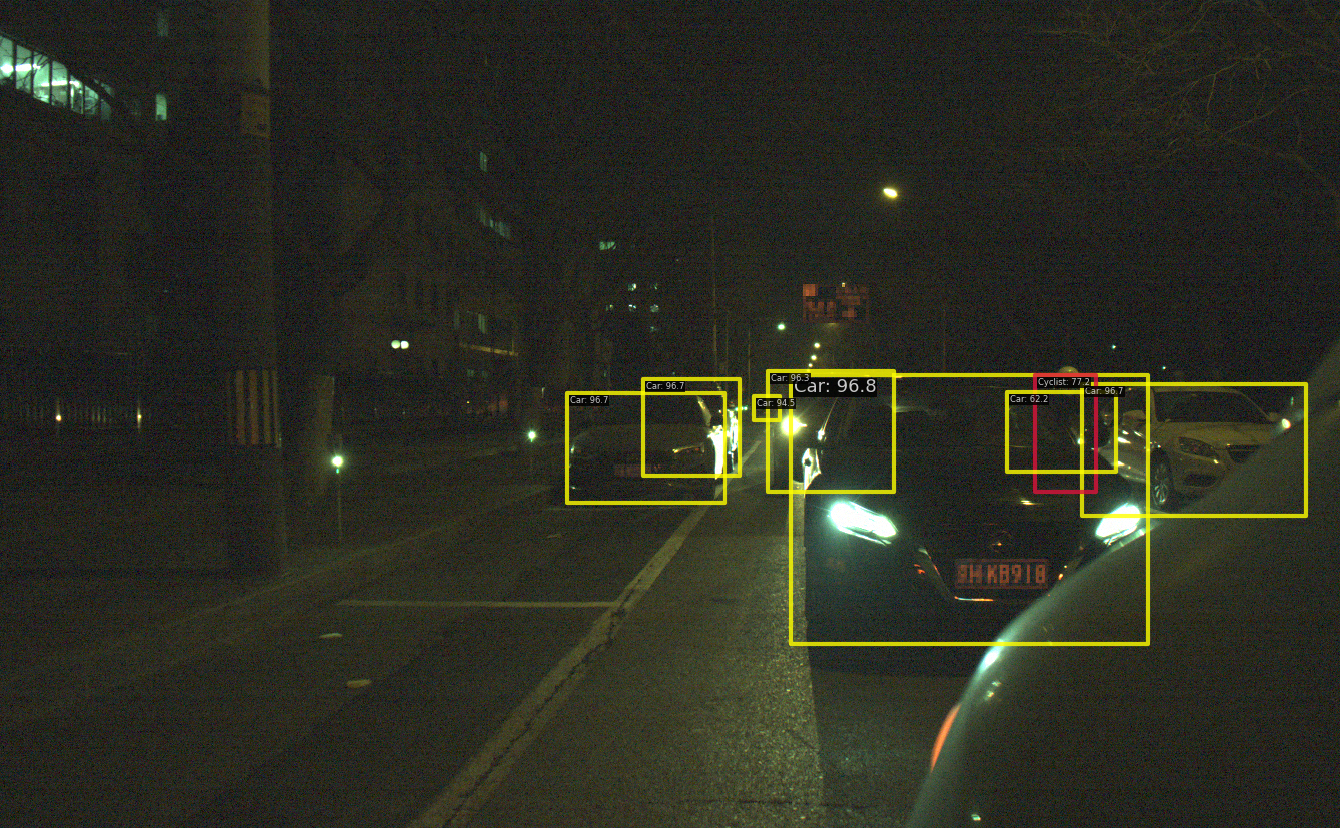} &
        \includegraphics[width=0.98\linewidth]{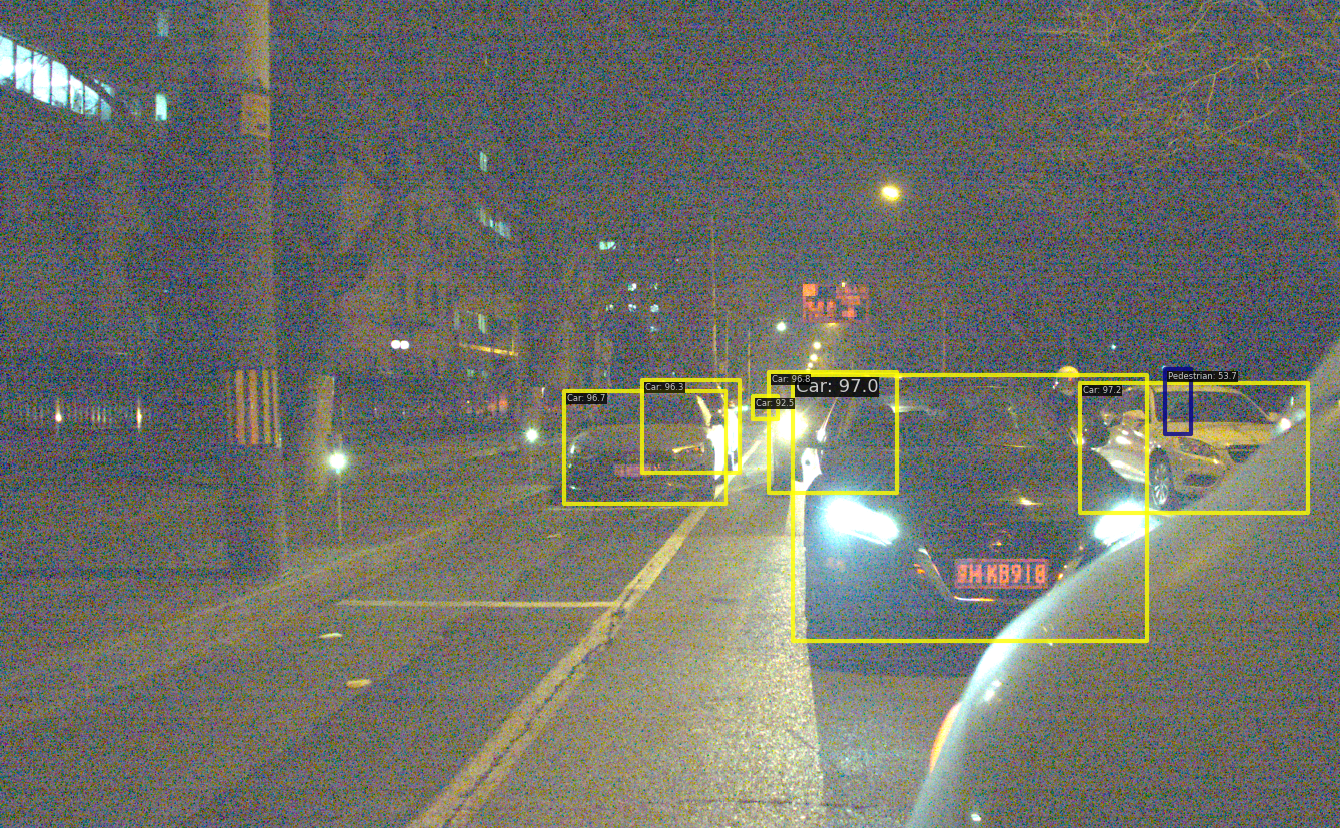} &
        \includegraphics[width=0.98\linewidth]{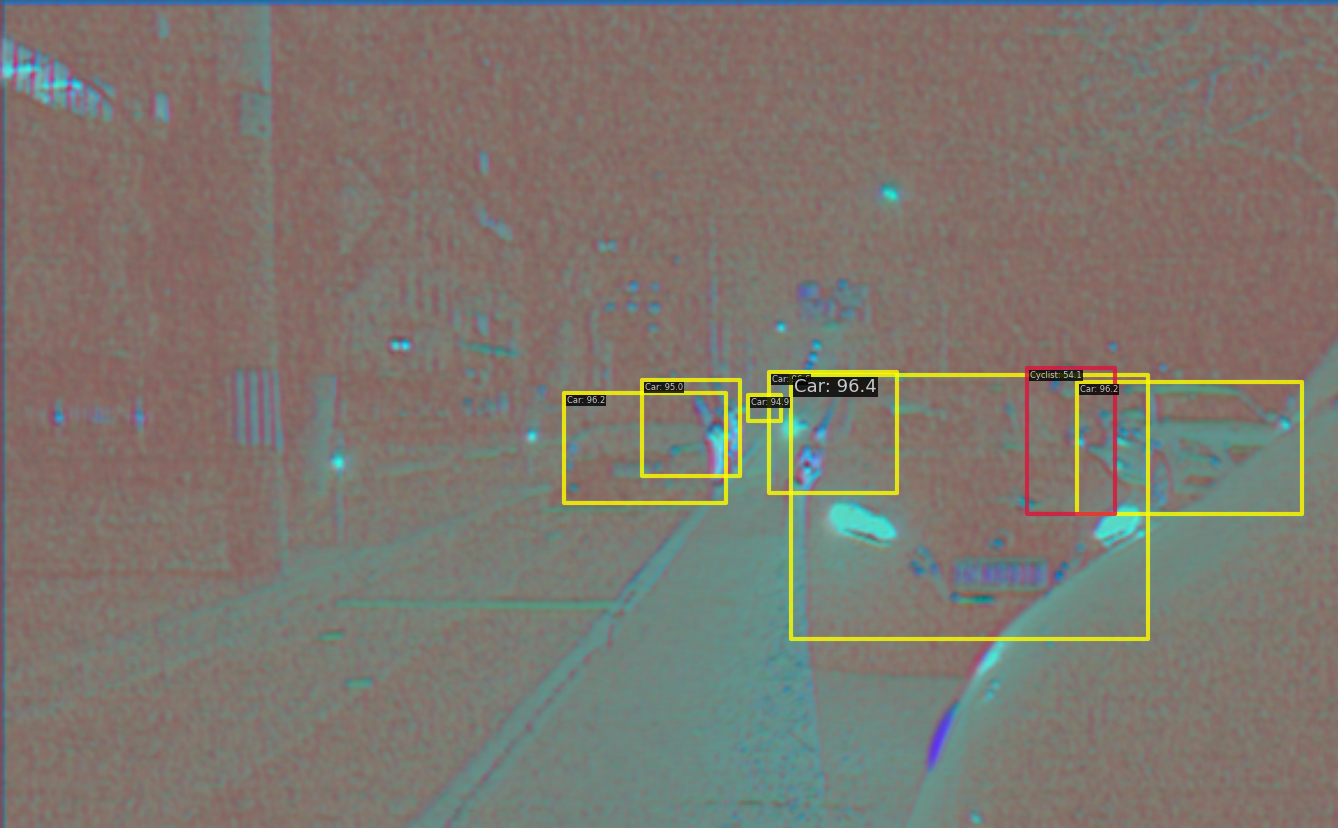} \\[2pt]

        \rotatebox{90}{\sffamily Denoised} &
        \includegraphics[width=0.98\linewidth]{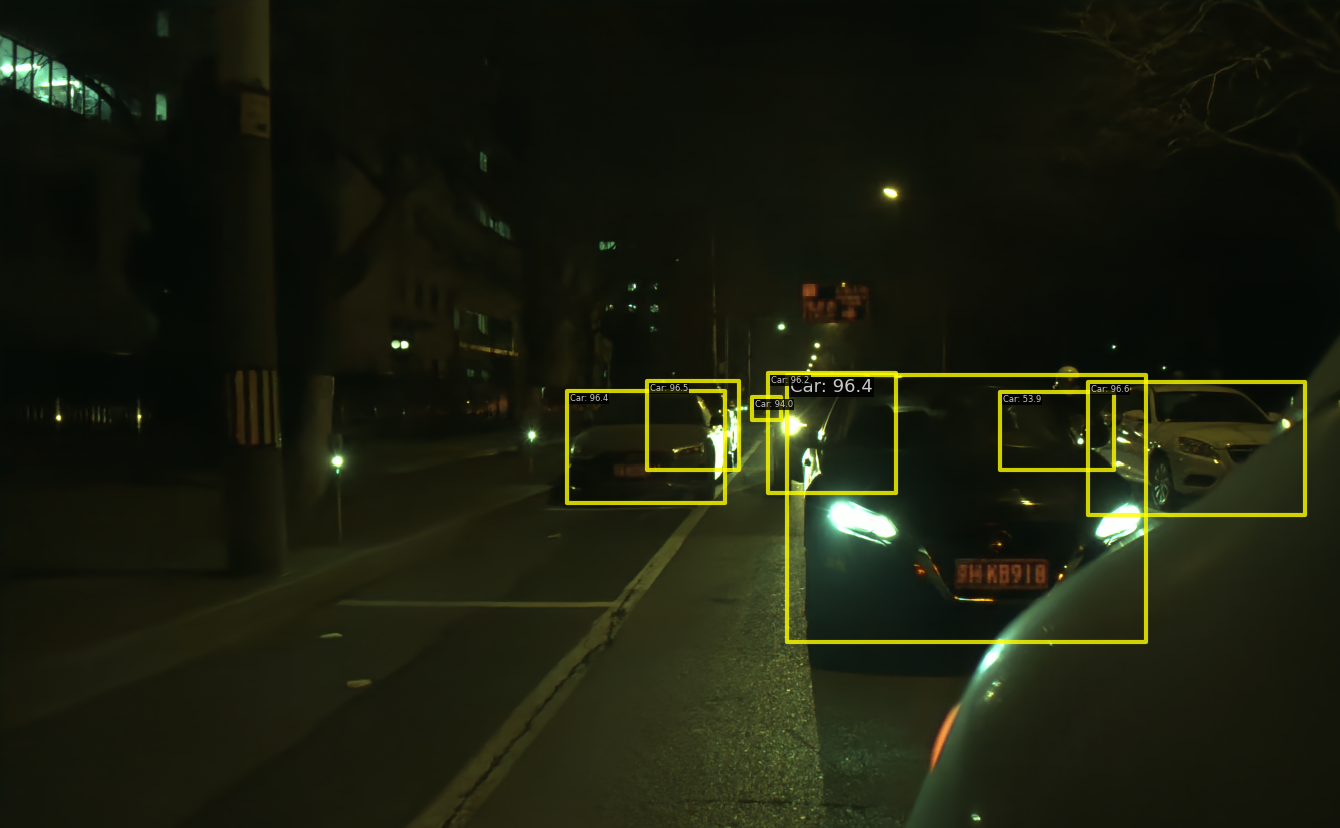} &
        \includegraphics[width=0.98\linewidth]{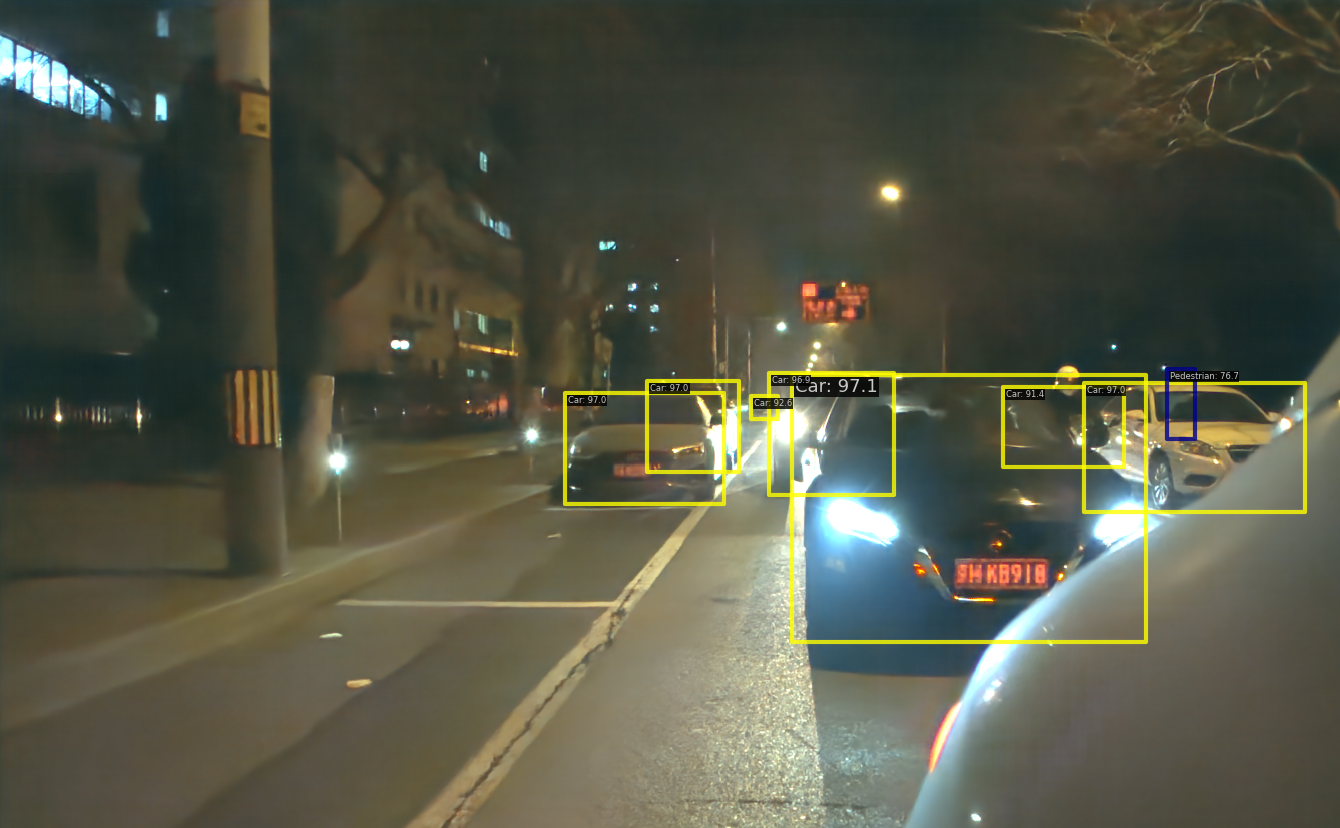} &
        \includegraphics[width=0.98\linewidth]{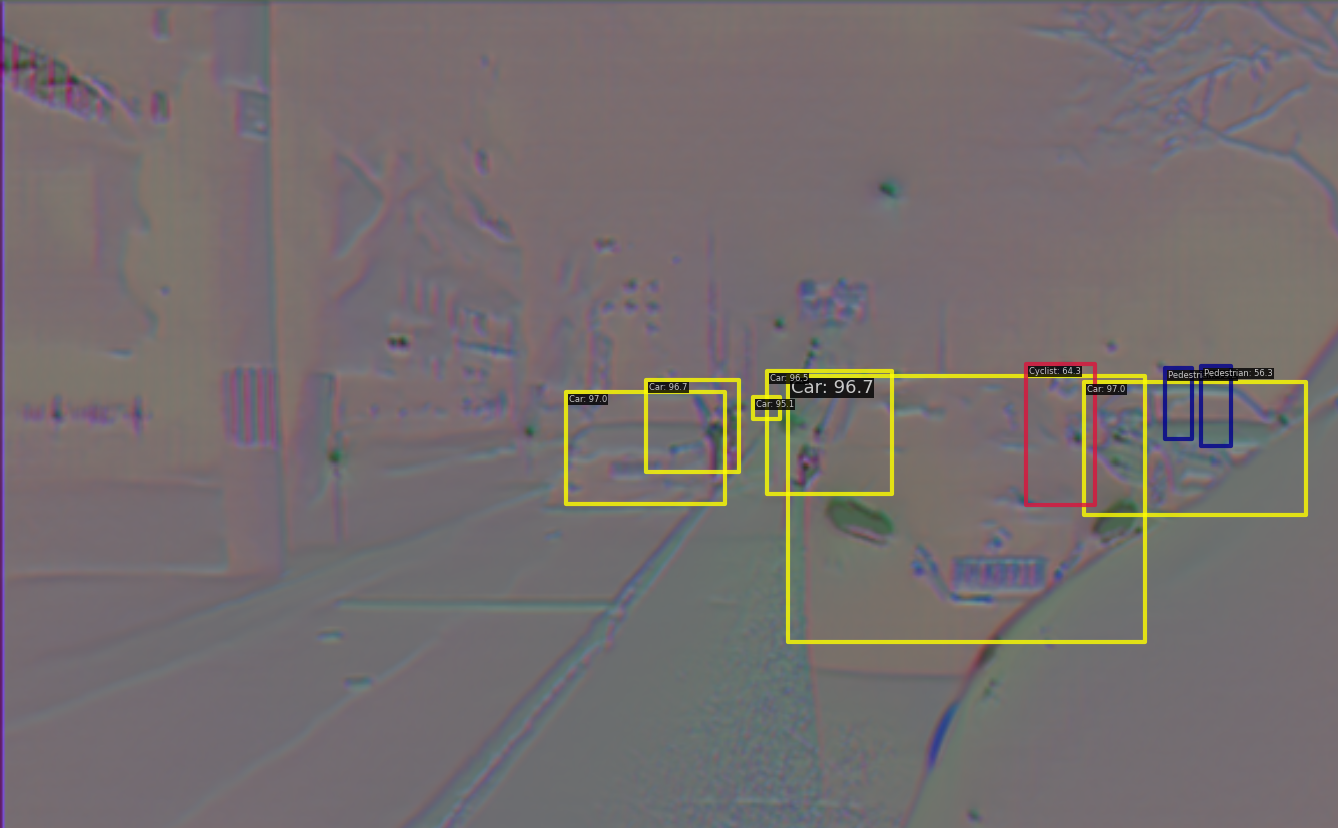} \\[5pt]
        
        & \multicolumn{1}{c}{{\sffamily RAW}} & \multicolumn{1}{c}{{\sffamily sRGB}} & \multicolumn{1}{c}{{\sffamily RAM (ours)}} \\
        
    \end{tabular}

    \caption{Qualitative comparison of images across RAW, sRGB, and RAM representations. The top row shows the ground truth (GT) image. The second row presents results on the noisy images, while the third row shows results on the denoised images.}
    \label{fig:noisy_denoised_comparison}
\end{figure*}

\cref{fig:noisy_denoised_comparison} presents detection results on noisy synthetic images and their denoised versions, as discussed in section 4.5.1. The presence of noise makes it challenging to detect occluded pedestrians and cyclists, including for RAM, as its output remains noisy without any specialized denoising component. The denoised images, produced by the state-of-the-art LED model \cite{jin2023lighting}, do not perfectly restore the original images and may introduce artifacts, which in some cases lead to false positives. In comparison, the best results appear in the bottom right image, where RAM, trained on denoised data, remains unaffected by any potential artifacts from the denoising process.

In conclusion, these visualizations emphasize the robustness of RAM in extreme conditions. Furthermore, they illustrate how RAM interprets images, often disregarding unnecessary or distracting features to focus on critical object features. Notably, we demonstrate these results on driving scenes, where accurate object identification in challenging conditions is essential and can be life-saving.

\section{Limitations}
While our proposed method demonstrates strong performance across a variety of datasets, it relies on a specific set of commonly used ISP functions tailored to these datasets. We are aware that these functions may not be suitable for all datasets, and other ISP functions might be necessary for different cases. However, the flexible design of RAM allows for easy addition or removal of ISP functions as needed.

Another challenge of working with RAW data is its sensor-specific nature. Unlike sRGB, where the ISP normalizes images to a standard 8-bit range, RAW images vary significantly in dynamic range, distribution, and characteristics across different sensors. This variability makes it more difficult to generalize a model trained on one sensor’s data to another compared to sRGB, where cross-dataset generalization is more straightforward. However, as shown in \cref{generalized_ram}, this limitation can be addressed by training a robust model on diverse sensor data, enabling better adaptability across different hardware.

\end{document}